# A Systematic Review of Deep Graph Neural Networks: Challenges, Classification, Architectures, Applications & Potential Utility in Bioinformatics

Adil Mudasir Malla, *IUST Kashmir, J&K( India)*   Asif Ali Banka, *IUST Kashmir, J&K.*

*Abstract*— In recent years, tasks of machine learning ranging from image processing & audio/video analysis to natural language understanding have been transformed by deep learning. The data content in all these scenarios are expressed via Euclidean space. However, a considerable amount of application data is structured in non-Euclidean space and is expressed as graphs, e.g. dealing with complicated interactions & object interdependencies. Modelling physical systems, learning molecular signatures, identifying protein interactions and predicting diseases involve utilising a model that can adapt from graph data. Graph neural networks (GNNs), specified as artificial-neural models, employ message transmission between graph nodes to represent graph dependencies and are primarily used in the non-Euclidean domain. Variants of GNN like Graph Recurrent Networks (GRN), Graph Auto Encoder (GAE), Graph Convolution Networks (GCN), Graph Adversarial Methods & Graph Reinforcement learning have exhibited breakthrough productivity on a wide range of tasks, especially in the field of bioinformatics, in recent years as a result of the rapid collection of biological network data.

Apart from presenting all existing GNN models, mathematical analysis and comparison of the variants of all types of GNN have been highlighted in this survey. Graph neural networks are investigated for their potential real-world applications in various fields, focusing on Bioinformatics. Furthermore, resources for evaluating graph neural network models and accessing open-source code & benchmark data sets are included. Ultimately, we provide some (seven) proposals for future research in this rapidly evolving domain. GNNs have the potential to be an excellent tool for solving a wide range of biological challenges in bioinformatics research, as they are best represented as connected complex graphs.

## I. INTRODUCTION

A graph specifies a data structure that defines a set of nodes & their interrelations. They are observed everywhere from social networks [1] to physical interactions [2]. Graphs can also represent unimaginable structures such as atoms, molecules, ecosystems, living organisms, planetary systems [3], the Internet (one giant graph) & more. Thus, the structure of the graph is found in our environment & perception of the world. It includes entities & their relationships to establish concepts such as reasoning, communication & relationships. Structured graph representation in these environments is expected; therefore, efficient & novel techniques are needed to solve graph-based tasks that would prove better than existing Machine Learning Models in dealing with non-Euclidean data & its generalisation concepts.

Deep learning gained considerable success in machine learning applications like classification & analysis of multimedia data in recent years. It is learning representations from crude data & making predictions via the learned representation, concurrently done by using various non-linear transformations (done in layers), training such models end to end by using learning methods based on gradient descent & efficiently solving learning problems. However, most deep learning theories revolve around interpreting conventional Euclidean data (Fig. 1a) but lag in non- Euclidean data where Graph Neural Networks come to the rescue. Deep learning also lacks relational & causal reasoning, abstraction, visualisation of intelligence & many other human capacities. With the growing amount of non-euclidean data captured via graph-structured data (Fig. 1b), many researchers have begun to devote more attention to treating graph-structured data that may express complicated interactions between objects.

The structure of calculations & representations in deep neural networks (DNNs) as graphs, typically known as Graph Neural Networks (GNN), is one of the ways to approach these problems. In general, GNN is an interconnectivity model that encapsulates graph interconnections by transferring messages across nodes, although taking into consideration the volume, diversity, & extensive topological information of the input data into the consideration. Currently, GNNs exhibit reliable efficacy in retrieving in-depth topological, i.e. structural information, and features from data & performing fast processing of big data like predicting properties of chemical molecules, perfecting missing networks [4], predicting drug interactions [5] & through user-product interactions to make highly accurate recommendations. The complexities of graph data put major obstacles to conventional machine learning algorithms since graphs can have an irregular structure, variable sizes of unordered nodes & different numbers of adjacent nodes of a given node. Image operations in space, like convolutions, are easier to calculate but are more complicated when applied to graph domains. In addition, the fundamental concept of contemporary algorithms of machine learning is that the objects are autonomous from each other. This presumption needs to be followed in graph data since different connections represent each object.

### A. Our Contribution

Our paper offers the following significant contributions:

**Comprehensive Survey:** We present the most extensive mathematical study of current deep learning algorithms for graph data to best of our knowledge apart from extensive

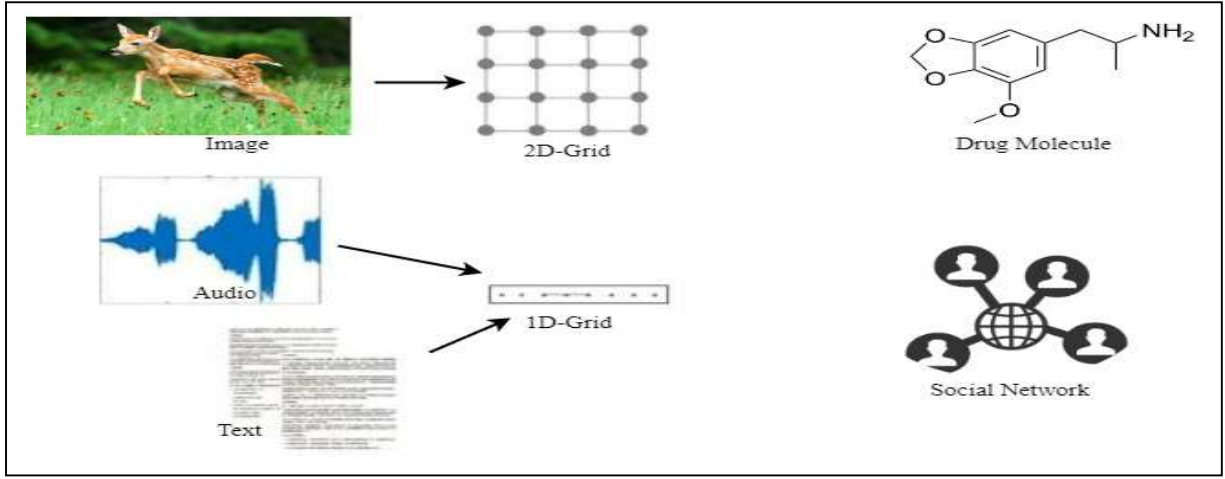

**Fig. 1** Examples of Euclidean & Non-Euclidean Data

(a) Euclidean Data  (b) Non- Euclidean Data

descriptions of representative models for each type of graph neural network.

**Resources in abundance:** This survey brings together a wealth of information on cutting-edge Graph neural networks, most of their practical applications, benchmark data sets, open-source code web links for hands-on guide to learning, utilising & developing various graph-based algorithms of deep learning, a for a variety of real-world applications

**Future Research Directions:** We analyse the limitations of existing approaches & propose seven potential future research tracks, especially highlighting model depth, scalability trade-off etc.

*B. The organisation of the Article*

The remaining sections of this survey are structured as follows Section **II** introduces, to the best of our understanding,

panorama of graph neural networks, various math notations & defines graph-related terms, including the general design principle of GNN. The Classification Taxonomy of graph neural networks is clarified in Section **III**. Graph Neural Network Models are discussed in sections **IV-VII**. A diverse collection of applications of GNNs from various disciplines is shown in Section **VIII.** Particularly, Bio-Informatics related applications of GNNs are described in section **IX.** Section X is related to Implementation & the current issues, and future directions are discussed in Section **XI.**

II. PRELIMINARIES & NOTATIONS

*A. Background*

With the advancement in processor technology in terms of speed & accuracy of processing data, Deep Neural Networks led to the development of Graph Neural Networks. The different milestones of GNNs are depicted in **Fig. 3**. Many in-depth analyses of graph neural networks have been published. Several recent surveys, like [6], [7], [8], concentrating on certain graph learning domains have also been conducted. However, our research takes a new approach to them, focusing primarily on both classic & advanced GNN models. In addition, we outline GNN versions /models along with their mathematical detailed mathematical descriptions for various graph types & provide a complete overview of GNN applications in most real-life sectors, particularly in Bio-informatics.

*B. Notations & Definitions*

Table I. defines various symbols & notations used in this paper, particularly for expressing various mathematical concepts and expressions. We use bold uppercase characters to signify matrices & bold lowercase characters to denote vectors throughout the literature work.

Table I NOTATIONS

| Notations | Descriptions |
|---|---|
| $G$ | A graph. |
| $N$ | set of vertices/nodes of graph |
| $v$ | A node $v \in V$. |
| $E$ | Edge set of graph |
| $e_{ij}$ | Edge $e_{ij} \in E$. |
| $N(v)$ | Neighboring nodes of $v$. |
| $A$ | adjacency matrix of $G$ |
| $A^T$ | Transpose of matrix $A$. |
| $A^n, n \in Z$ | nth power of adjacency matrix. |
| $[A, B]$ | Concatenation of $A$ & $B$. |
| $\|\cdot\|$ | Length of a set. |
| $\odot$ | Element-wise product. |
| $n$ | Total count of nodes, $n = \|V\|$. |
| $m$ | Total count of edges, $m = \|E\|$. |
| $d$ | node feature vector dimension. |
| $b$ | hidden node feature vector dimension. |
| $c$ | edge feature vector dimension. |
| $k$ | Layer index |
| $t$ | Time step/iteration index |
| $D$ | Degree matrix of $A_{ij}$, $D_{ii} \sum_{j=1}^{n} A_{ij}$ |
| $X \in R^{n \times d}$ | Graph feature matrix |
| $x \in R^n$ | Graph feature vector with $d = 1$ |
| $x_v \in R^d$ | Feature vector of $v$ |
| $X^e \in R^{m \times c}$ | feature matrix of edge $e$. |
| $x^e_{(v,u)}$ | Edge, $e(v, u)$ feature-vector |
| $X^{(t)} \in R^{n \times d}$ | Feature-matrix of graph at time $t$. |
| $H \in R^{n \times b}$ | Hidden node feature matrix. |
| $h_v \in R^b$ | Hidden node $v$ feature vector. |
| $W, \Theta, w, \theta$ | Learnable model parameters. |
| $\sigma()$ | Sigmoid threshold function. |
| $\sigma_h()$ | Tangent hyperbolic threshold function. |

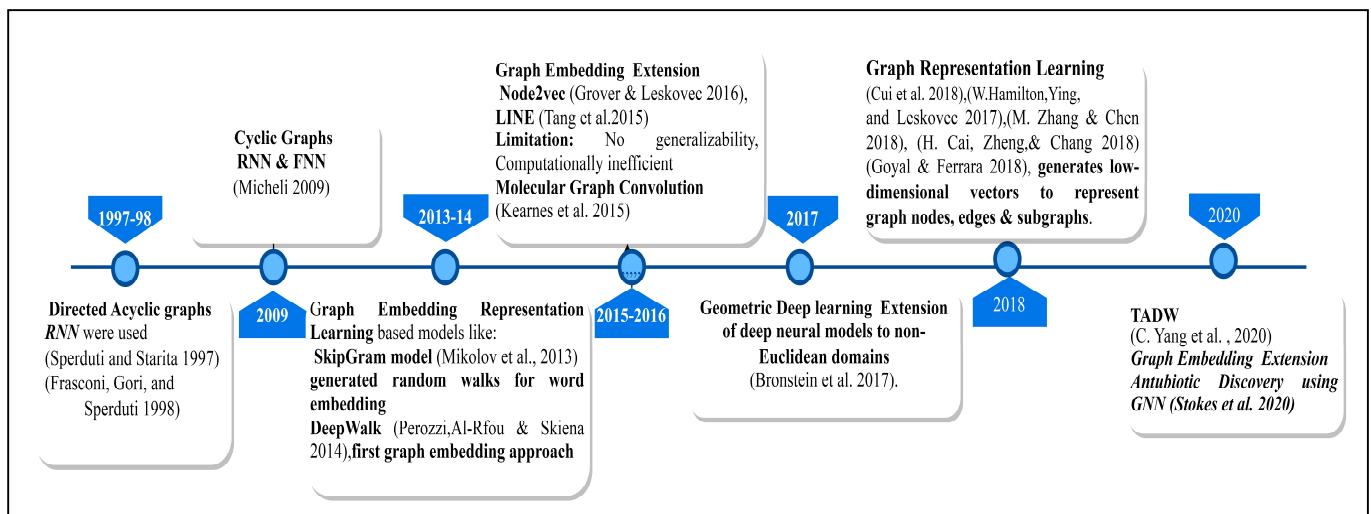

**Fig. 3.** Milestones of GNN

**Graph:** *G = (E, V)* symbolises a graph, with a set of vertices *V* connected via another set of edges *E*.

**Directed/Undirected Graph:** A graph in which every edge between nodes is directed is termed a directed graph. Otherwise, a graph is undirected.

**Static/ Dynamic Graphs:** A spatial-temporal or Dynamic graph is an attributable graph where its node features change dynamically over time & Static Graphs are contrary to dynamic graphs for feature change with time. The spatial-temporal graph symbolised as $G^{(t)} = (V, E, X^{(t)})$ is an attributed graph with dynamically changing node properties throughout time. $G^{(t)} = (X^{(t)}, E, V)$, $X^{(t)} \in R^{n \times d}$ is the spatial-temporal graph.

**Homogeneous/Heterogeneous Graphs:** Edges & vertices in homogeneous graphs are of the same types, while they are of varied types in heterogeneous graphs.

Signed Graphs The graphs consist of both positive & negative edges, e.g. Online social networks where users may block or unfollow (denoted as negative edges[9] other users.

**Hypergraphs:** The graphs which consist of more than two nodes connected by only an edge are known as hypergraphs, e.g. Citation networks.

**Large Graphs:** When the size of the graph is such that its adjacency matrix is of the order of $O(n^2)$ for its space complexity, they are treated as large graphs.

Generally, the tasks/Roles for learning a deep neural network model on graphs are divided into three groups as defined below:

**Node/vertex-level operations** include tasks like vertex classification, regression and clustering. Node categorisation tries categorising nodes into various classes, while vertex regression recognises a continuous value for each vertex. Vertex clustering proposes differentiating the edges into many disjoint segments, where similar nodes should be in the same segment, e.g. node recommendation & Citation network.

**Edge level operations** edge/connection classification or prediction. The model requires categorising the edge or detecting if an edge exists between two specific nodes, e.g. predicting side-effects of drugs.

**Graph-level tasks** involve graph-level classification, regression & matching. All this necessitates the model to learn the graph structure. Examples include image classification, Enzyme Detection among the set of proteins.

All the tasks for learning the deep graph models are done under three basic training setting modes, which are defined as under:

The **supervised settings** provide labelled data for training. The **semi-supervised configuration**s have a large set of unlabelled /untagged nodes & small set of tagged nodes for training. **The inductive setting** demands the model to predict the label of a specific unlabeled node during the testing phase. In contrast, the **transductive/derivative setting** provides a new unlabeled node from the same distribution for inference. Most node & edge classification tasks are semi-supervised. Most recently, [7], [5] proposed a mixed inductive-inductive scheme, longing for a new way to mixed settings**. The unsupervised setting** provides the model with only unlabeled data to find the pattern. Node clustering is a typical unsupervised learning task. A specific loss function can be designed for the task given the task type & training settings. For instance, for node-based semi-supervised classification tasks,cross-entropy loss of marked nodes is used in the training set.

*C.    GNN Model General Design Principle*

In general, the principle design pipeline of GNN has four steps [10] which are defined as:

**i) Identify graph structure:** First, find out the graph structure that fits the problem definition. In structural contexts, the graph structure is evident inside the applications, e.g. physical systems, molecular applications and knowledge-based graphs. In non-structural applications, graphs are implicit to the corresponding task, e.g. developing a "word" graph that is fully connected to text or designing the image scene graph.

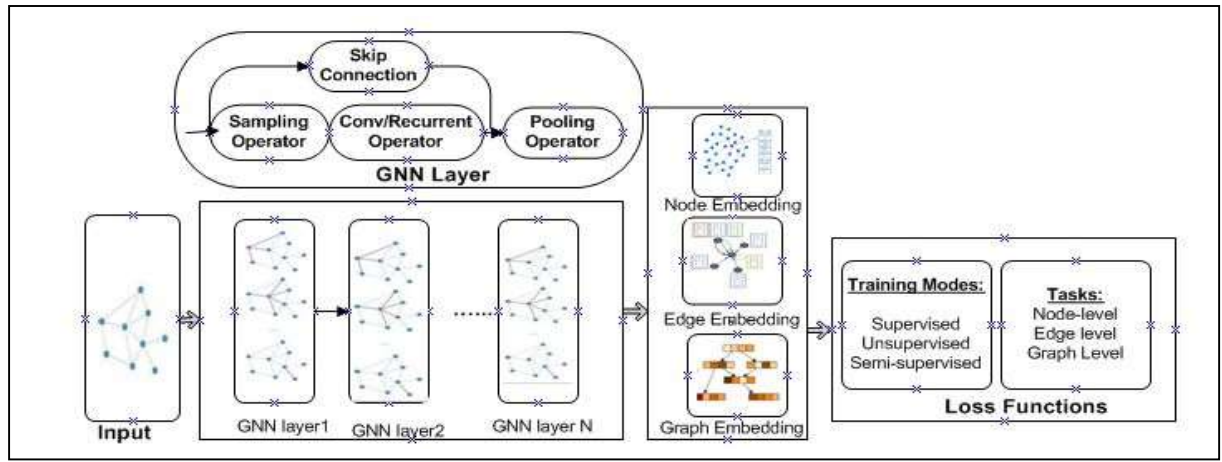

1. Find structure
2. Specify type/scale
4. Build model using computation modules
3. Design loss functions

**Fig. 4.** A GNN Model—General Pipeline Design

**ii) Specify graph type/scale:** Graphs with complex types can provide more information on vertices & their edges. Various types of graphs are already defined in the previous section. There is no definitive distinction between "small" and "large" graphs considering the scale of graphs. With the development of computing technology, the standards continue to evolve (like the memory & speed of GPUs).

**iii) Error function Design**: The error function design is based on the training setting & type of task. There are commonly three types of tasks [11], as already mentioned above, regarding deep graph model learning challenges. Similar to traditional deep learning models, the learning step of GNNs entails employing gradient-based refinement to minimise a loss/error function (e.g., [12]. It should be noted that the cross entropy (CE) error/loss function is often employed for classification tasks. There are various other alternatives, including hinge-losses, mean square-error (MSE) & mean absolute error (MAE). The CE is an intuitive choice because of its close link to traditional logistic & multinomial regression models. The general architecture of the model of **GNN** is depicted inside the midst of above Fig. 4.

**iv) Build model using computation modules**

Computation modules in GNN are mathematical computations executed to either propagate & summarise information from neighbouring nodes or are designed to reduce the feature vector sizes & eliminate their corresponding redundancy at a node, edge and graph-level prediction tasks. Computation modules based on different high-level operations they are performing are divided as:

**a) Propagation module:** They propagate information between nodes, allowing aggregation of information from neighbours using convolution & recurrent operators to capture both functional/feature & topology information. In contrast, skip connections operations collect information from a node's historical representation & mitigate the over-smoothing problem.

**b) Sampling module:** Larger graphs usually require a sampling module to propagate information to a subset of each node's neighbours in a giant graph to avoid neighbours exploding problems. This is the reason why sampling modules are usually combined with propagation modules.

**c) Pooling module**: To condense the node's features & produce higher-level features that can capture the details of the entire graph or a subgraph for graph-level tasks, the graph pooling layer/module is used. A standard GNN model is generally developed by stacking these computing units.

## II. TAXONOMICAL CLASSIFICATION OF GRAPH NEURAL NETWORK

GNNs represent a subclass of Deep Learning Procedures intended to execute inference on graph-based data. GNNs are applied to graph data and can carry out prediction tasks at the node, edge, and graph levels. Apart from introducing the broad classification of Graph Neural Networks based on different training settings or architecture, as shown in Fig. 5, this section describes the GNN Types based on computational, i.e. calculation modules, which are at the heart of any Graph Neural Network. The broad categorisation of Graph Neural Networks is briefly characterised in Table II and is described in detail in upcoming sections.

Regarding the computation unit operators-- the propagation units, sampling units & pooling units constitute basic computation modules under various training settings, i.e. semi-supervised, supervised & unsupervised learning settings of GNNs as depicted in Fig. 6.

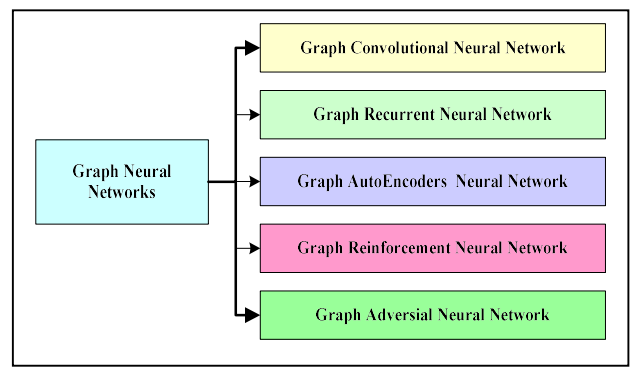

Fig. 5 Broad Classification of GNNs

TABLE II. COMPARISON OF TYPES OF GNNs

| S.No. | Category | Main Aims/Assumptions | Primary Functions |
|---|---|---|---|
| 1. | Graph Convolution Networks | Common graph structure patterns at both the local and global levels | Graph-convolution and readout operations |
| 2. | Graph Recurrent Neural Networks | Sequential & Recursive patterns of the graphs | States of nodes and graphs are defined. |
| 3. | Graph Reinforcement learning | Constraints & Feedbacks of the graph-tasks | Actions/rewards based on the graph |
| 4. | Graph Auto encoders | Low-rank structures of graphs | Learning the node representation Independently in unsupervised mode |

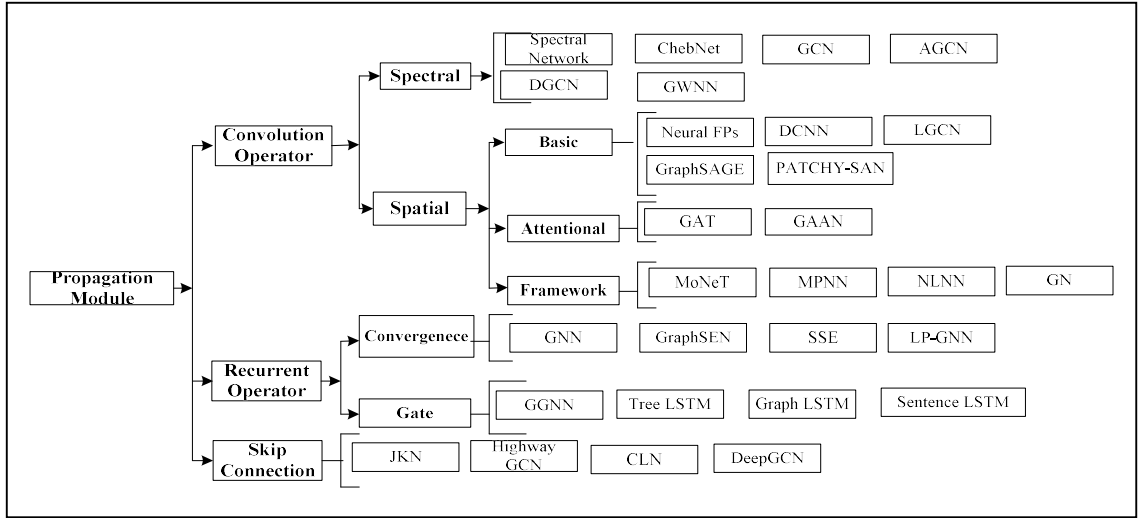

Fig. 6. Propagation Modules based on different operators, methods & frameworks

## III. BASIC PROPAGATION MODULE--- GRAPH NEURAL NETWORKS

The propagation unit is a computing module employed to transmit information across nodes such that the aggregated information obtained from neighbouring nodes can capture both features topological/structural In propagation units, the recurrent operator & the convolution operator are used to gather data from neighbours whilst skip connection operations are utilised to acquire data from the past representation of nodes to minimise the over-smoothing condition.Fig.4 below shows the main Propagation Modules based on different operators, methods & frameworks used for propagating the information leading to the classification of GNNs for different applications.

### A. Propagation module - Convolution Operator

Convolution operators are the most widely used propagating operators of GNN models. The fundamental idea behind the operator of the graph convolution is to extend convolutions from other domains, i.e. grid data, to the graph domain. The graph convolution module characterises node $v$ via aggregating its own attributes $x_v$ & of its neighbour's $x_u$, whereby $u$ belongs to adjacent nodes $N(v)$ connected in an irregular fashion. Such GNNs are known as Convolution Graph Neural Networks (ConvGNNs). They overlay several graph convolution layers to get high-level node descriptions. A ConvGNN for node & graph classification is shown in Fig. 7a & Fig. 7b.Many complex GNN models rely on ConvGNNs for their construction and are further classified as spectral & spatial methods.

### 1) Spectral Convolution Approach

Theoretically, these approaches are built on the graph signal processing concept [13] & define convolution operator in the spectral domain. Since GCN [14] closed the gap between spectrum-based & space-based techniques, Spatial approaches have evolved tremendously due to their appealing efficiency, universality, and adaptability. In spectral-CNN techniques, firstly, the graph-signal **x** is projected into the space domain using Fourier graph transform, and then an operation of convolution is executed. Finally, the resulting signal is converted back via the operation of inverse Fourier graph transform $\mathcal{F}^{-1}$. Such transformations are stated as follows:

$$(\mathbf{x}) = \mathbf{U}^T \mathbf{x} \quad (1)$$
$$\mathcal{F}^{-1}(\hat{\mathbf{x}}) = \mathbf{U}\hat{\mathbf{x}} \quad (2)$$
$$\mathbf{x} = \sum_i \hat{x}_i \mathbf{u}_i \quad (3)$$

wherein **U** denotes eigenvector matrix of normalised Garph-Laplacian stated as $L = I_n - D^{-(1/2)} A D^{-(1/2)}$. $A$ denotes the adjacency matrix & $D$ denotes the degree matrix of a graph. As the adjusted laplacian of the graph is

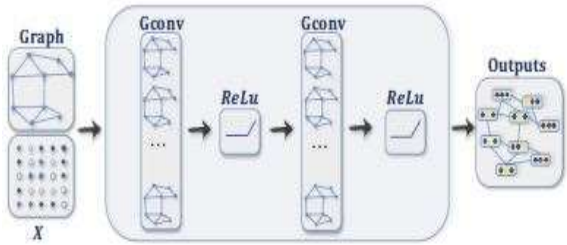

(a) **Classification of Node:** Multiple graph-convolution layers of ConvGNN. Every layer generates latent representation for each vertex via feature-data aggregation from its neighbours but then applies transfornation in non-linear fashion on resulting outputs.

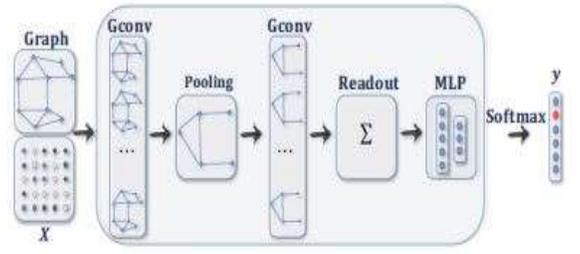

(b) **Classification of Graph:** Readout or pooling layers of **ConvGNN** coarsens the graph to sub-graphs inorder to depict higher-level representation of graphs, then summarising the ultimate graph graph using the mean/sum of latent depictions of sub-graphs.

**Fig. 7.** ConvGNN for Node & Graph Classification

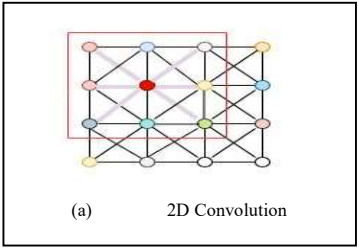

(a) 2D Convolution

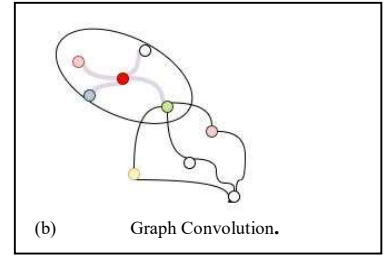

(b) Graph Convolution.

**Fig. 8.** 2D Convolution vs. Graph Convolution

symmetric positive semi-definite real in nature, it is factorised as $L = U\Lambda U^T$, wherein $\Lambda$ is a diagonal matrix representing eigenvalues. Following on convolution theorem [15], the convolution operation is expressed as:

$$\mathbf{x} * G\mathbf{g} = \mathcal{F}^{-1}(\mathcal{F}(\mathbf{x}) \odot \mathcal{F}(\mathbf{g}))$$
$$= \mathbf{U}(\mathbf{U}^T\mathbf{x} \odot \mathbf{U}^T\mathbf{g}) \quad (4)$$

Where $\mathbf{U}^T\mathbf{g}$ is filtered in spectral domain & $\odot$ denotes elementwise dot product. If we simplify the filter by using a learnable diagonal matrix $\mathbf{g}_\theta = \mathrm{diag}(\mathbf{U}^T\mathbf{g})$, then the spectral graph is simplified as:

$$\mathbf{X} * G\mathbf{g}_\theta = \mathbf{U}\mathbf{g}_\theta \mathbf{U}^T\mathbf{x} \quad (5)$$

Every spectral-based ConvGNNs suits this specification, as the basic difference relies on choices of filter $\mathbf{g}_\theta$. Table III. introduce several typical spectral methods based on different filter*s and* $\mathbf{g}_\theta$ designs.

Studies have recently explored other symmetric matrices to make incremental improvements over GCN [16] Adaptive Graph Convolution Network (AGCN) [17] learns latent structural relations that the graph adjacency matrix does not specify. All of these models employ the original graph structure to represent node relationships. However, there may be hidden relationships between nodes. An adaptive Graph Convolution Network (AGCN) is proposed [18] to discover underlying relationships. A "residual" graph laplacian is learned by AGCN & added to the original Laplacian matrix. As a result, it works for diverse graph-structured datasets.

### 2) Spatial Convolution Approach

Approach where convolutions are defined directly on graph topology. The most challenging aspect of this approach is specifying convolution with different-sized neighbourhoods while ensuring CNN local invariance.

**a)** *Basic Spatial Convolution Approach*

Spatial-based techniques define graph convolutions relying on spatial relations of nodes, similar to the convolution operation of a standard CNN on an image. Images can all be a graph representation wherein every pixel symbolises a node. In 2D convolution, neighbours of each pixel in an image is determined by the filter size and considered nodes of a graph. As shown in **Fig. 8a**, an arithmetic mean of the pixels of the red vertex along with its neighbours is employed & note that the neighbourhood of a node is ordered & fixed in size., Spatial-based graph convolutions in Fig. 8b interpolate the middle red node's depiction by taking the average value of its features with that of its neighbours to produce an updated hidden representation for the centre node where, in contrast to image data, a node's neighbours are unordered and vary in size. Spatial-based ConvGNNs & RecGNNs have the same principle of information propagation & message transfer. The convolution spatial graph technique propagates node information along edges. Various models studied under spatial-based ConvGNNs are:

i) **Neural Network for Graphs** (NN4G) [19] which is proposed in parallel with GNN. NN4G learns graph mutual dependency through a compositional neural architecture with distinct parameters at each layer, which sets it apart from RecGNNs. Through progressive architectural construction, node neighbourhoods can be scaled-up. NN4G conducts graph convolutions via simple summation of the node's neighbourhood information & further employs skip & residual connections for retaining information along each layer. NN4G estimates node states at the next layer as a byproduct defined below:

$$\mathbf{h}_v^{(k)} = f\left(\mathbf{W}^{(k)^T}\mathbf{x}_v + \sum_{i=1}^{k-1}\sum_{u \in N(v)} \boldsymbol{\Theta}^{(k)^T}\mathbf{h}_u^{(k-1)}\right) \quad (6)$$

where $h_v^{(0)} = 0$ & $f(.)$ is an activation function. Equation (6) can be denoted in matrix form as:

$$\mathbf{H}^{(k)} = f\left(\mathbf{X}\mathbf{W}^{(k)} + \sum_{i=1}^{k-1} \mathbf{A}\mathbf{H}^{(k-1)}\boldsymbol{\Theta}^{(k)}\right) \quad (7)$$

This is similar to GCN [20] except for using an un-normalised adjacency matrix which could result in significantly different scales across hidden node states in NN4G. Contextual Graph Markov Model (CGMM) [21] is the probabilistic method that relies on NN4G.CGMM provides the advantage of probabilistic interpretability while maintaining spatial locality.

**ii) NeuralFPs** According to [3], neural FPs employs several weight matrices for nodes of various degrees where the weight matrix for nodes at layer **t+1** with degree $|\mathcal{N}_v|$ is represented by $W^{t+1}_{|\mathcal{N}_v|}$ as expressed below:

$$t = h_v^t + \sum_{u \in \mathcal{A}_v} h_u^t \qquad (8)$$
$$h_v^{t+1} = \sigma(tW^{t+1}_{|\mathcal{N}_v|}) \qquad (9)$$

Hence, it is not applicable to large-scale networks with large node degrees. However, Neural FPs learnt parameters each layer. NN4G estimates node states at next layer as a byproduct, defined below as:

$$h_v^{(k)} = f(W^{(k)^T}x_v + \sum_{i=1}^{k-1}\sum_{u \in N(v)} \Theta^{(k)^T} h_u^{(k-1)}) \qquad (10)$$

where $h_v^{(0)} = 0$ & $f(.)$ is an activation function. Equation (6) can be denoted in matrix form as:

$$H^{(k)} = f(XW^{(k)} + \sum_{i=1}^{k-1} AH^{(k-1)}\Theta^{(k)}) \qquad (11)$$

This is similar to GCN [22] except use of an un-normalized adjacency matrix which could result in significantly different scales across hidden node states in NN4G.Contextual Graph Markov Model (CGMM) [21] is the probabilistic method that relies on NN4G.CGMM provides advantage of probabilistic interpretability while maintaining spatial locality.

**ii) NeuralFPs** According to [3], neural FPs employs several weight matrices for nodes of various degrees where weight matrix for nodes at layer **t+1** with degree $|\mathcal{N}_v|$ is represented by $W^{t+1}_{|\mathcal{N}_v|}$ as expressed below:

$$t = h_v^t + \sum_{u \in \mathcal{A}_v} h_u^t \qquad (12)$$
$$h_v^{t+1} = \sigma(tW^{t+1}_{|\mathcal{N}_v|}) \qquad (13)$$

Hence, it is not applicable to large-scale networks with large node degrees. But, Neural FPs learnt parameters of varied degree nodes, which correspond to discrete & inherent edge attributes of various bond types in the chemical graph, and thereafter added the outputs.

**iii) DCNN** Diffusion Graph convolutions are considered as process of diffusion executed via **Diffusion Convolution Neural Network** (DCNN) [23].Information is sent with a predetermined probability of transition from one node to its neighbours enabling its distribution to attain equilibrium after successive iterations. Diffusion graph convolution is expressed as $H^{(k)} = f(W^{(k)} \odot P^k X)$ where $f(.)$ denotes activation/threshold function & $P=D^{-1}A$ computes probability transition matrix $P \in R^{n*n}$. Hidden represented matrix $H^{(k)}$ in DCNN has identical dimensions as the feature- input matrix $X$ & is independent of prior latent representation matrix $H^{k-1}$.Regarding final model results, DCNN appends $H^{(1)}, H^{(2)},...,H^{(K)}$. Additionally, DCNN transforms each edge into a node connected with its tail and head. After undergoing this transition, edge features can be used as node features.

**iv) Diffusion Graph Convolution (DGC)** [20] adds up outcomes at every diffusion step rather than appending them as static-distribution of the diffusion procedure, it represents the total sum of the power series of transition-probability matrices. Thus overall it is redefined as :

$$H = \sum_{k=0}^{K} f(P^k X W^{(k)}) \qquad (14)$$

Where $W^{(k)} \in R^{D*F}$. **Probability matrix** based on **transition** stipulates that remote neighbours provide minimal information to central node.

**v) Parametric Graph Convolution (PGC-DGCNN)** [22] enhances influences of remote neighbors dependent on the Shortest routes. If the shortest-distance between vertex **v** to **u** is identical to length **j** determined via adjacency matrix $S^{(j)}$ of the shortest-routethen $S^{(j)}_{v,u} = 1$,else **0**.PGC-DGCNN as introduced below provides a graph convolution procedure using hyper - parameter r to modulate receptive size of field..

$$H^{(k)} = \|_{j=0}^{r} f\left((\tilde{D}^{(j)})^{-1} S^{(j)} H^{(k-1)} W^{(j,k)}\right) \qquad (15)$$

Where $\tilde{D}^{(j)}_{ii} = \sum_l S^{(j)}_{i,l}$, H(0) = X, & $\|$ denotes vector concatenation. Computing shortest route adjacency matrix can be time-consuming, $O(n^3)$. PGC [24] divides neighbors of a node into **Q** groups depending on multiple criterion that aren't restricted to shortest pathways. The PGC generates **Q** adjacency-matrices depending on each group's specified neighbourhood. After that PGC uses GCN [14]] with a distinct parameters matrix for every neighbour group & adds outputs as

$$H^{(k)} = \sum_{j=1}^{Q} \bar{A}^{(j)} H^{(k-1)} W^{(j,k)} \qquad (16)$$

Where $H^{(0)} = X$, $\bar{A}^{(j)} = (\tilde{D}^{(j)})^{-\frac{1}{2}} \tilde{A}^{(j)} (\tilde{D}^{(j)})^{-\frac{1}{2}}$, $\tilde{A}^{(j)} = A^{(j)} + I$. It is impractical to assume full size of a node-neighbourhood since number of neighbours can vary from one to a thousand or more.

**vi) PATCHY-SAN** strategy retrieves & normalises a neighborhood of precisely **k** number of nodes for each node [25]. Receptive field is normalized-neighborhood in a typical convolution operation.

**vii) LGCN** CNNs are also used as aggregators in **Learnable graph convolution network (LGCN)** [26]. It uses max pooling to retrieve top-k feature elements from node neighborhood matrices then uses 1D -CNN to compute hidden representations.

In order to include edge characteristics, LGCN designed a line graph. The Nodes of line-graph are represented by directed-edges as in original-graph. The two nodes in a line-graph are connected if data flows along the edge that connects them in the source-graph. Following this, an LGCN provided two separate GCNs for the source graph & its corresponding line graph.

**viii) GraphSAGE** [27] is a general inductive framework for generating embedding from a node's local neighborhood by sampling & aggregating features as mentioned below by uniformly sampling a certain number of neighbours instead of using entire neighbor set:

$$h^{t+1}_{\mathcal{N}_v} = AGG_{t+1}(\{h_u^t, \forall u \in \mathcal{N}_v\}) \qquad (17)$$
$$h_v^{t+1} = \sigma(W^{t+1} \cdot [h_v^t \| h^{t+1}_{\mathcal{N}_v}]) \qquad (18)$$

Wherein $AGG_{(t+1)}$ denotes aggregation function. The aggregators suggested by GraphSAGE are mean, LSTM,

& pooling aggregators. Mean aggregator-based GraphSAGE is viewed as the **inductive variant of the GCN**. However, the aggregator of LSTM & hence requires the defined order of nodes.

**ix) Relational GCNs (R-GCNs)** [28] trained different weights for various edge types (discrete edge features) of knowledge graphs.

**b) Spatial Attention-based Convolution Approach**

**i) Graph Attention Network (GAT)** [29] implies that the contribution of nearby nodes towards the central node of GAT is neither fixed like GCN [14] nor same as GraphSage [27]. Many sequence-based operations, such as machine reading, translation [30] & so on, have successfully employed attention mechanisms. Several models also attempt to generalise the attention mechanism on graphs [31],[32]. Attention-based operators, in contrast to previously mentioned operators, assign multiple weights to neighbours to reduce noise & obtain better results. GAT integrates attention mechanisms into the propagation step to learn weights in the relation between two adjacent vertices. GAT-graph convolution function specifies the hidden state of node *v* as:

$$\mathbf{h}_v^{(k)} = \sigma\left(\sum_{u \in \mathcal{N}(v) \cup v} \alpha_{vu}^{(k)} \mathbf{W}^{(k)} \mathbf{h}_u^{(k-1)}\right) \quad (19)$$

Where $h_v^{(0)} = x_v$ & **W** is the weight matrix related to linear transformation performed on each node. Associative strength between node *v* & its neighbour *u* is measured by attention weight, $\alpha_{vu}^{(k)}$:

$$\boldsymbol{\alpha}_{vu}^{(k)} = \textbf{softmax}\left(g\left(\mathbf{a}^T[\mathbf{W}^{(k)}\mathbf{h}_v^{(k-1)} \parallel \mathbf{W}^{(k)}\mathbf{h}_u^{(k-1)}]\right)\right) \quad (20)$$

Wherein *g(.)* denotes the LeakyReLU threshold function **& a** symbolises learnable parameters weight vector of single-layer MLP. Multi-headed attention is also performed via GAT to increase the model's expressibility. This is a significant refinement over GraphSage in node classification, whereas GAT supposes that all attention heads contribute equally.

**iii) GaAN** Multi-head attention mechanism is also used by gated attention network (GaAN) [32] to stabilise the learning process [30]. However, it replaces GAT's average operation with a self-attention mechanism that gathers input from various heads. GeniePath [6] includes a gating mechanism analogous to LSTMs to manage the flow of information via graph convolution layers simultaneously. Besides applying graph attention spatially. Other graph attention models include [33],[34] not included in ConvGNN architecture. Multi-head attention computes hidden states using **K**-independent attention head matrices and then concatenates/averages their features following two output representations:

$$\mathbf{h}_v^{t+1} = \parallel_{k=1}^{K} \sigma\left(\sum_{u \in \mathcal{N}_v} \alpha_{vu}^k \mathbf{W}_k \mathbf{h}_u^t\right) \quad (21)$$

$$\mathbf{h}_v^{t+1} = \sigma\left(\frac{1}{K}\sum_{k=1}^{K}\sum_{u \in \mathcal{N}_v} \alpha_{vu}^k \mathbf{W}_k \mathbf{h}_u^t\right) \mathbf{x} \quad (22)$$

The normalised attention coefficient of the kth attention head is denoted by $\alpha_{ij}^k$. This architecture computes node-neighbour pairs in parallel, employed on graph nodes of varying degrees by providing random weights to neighbours & readily to problems in inductive learning[35].

**c) General Frameworks for Spatial Approaches**
Several general frameworks are presented to integrate several models into a unified framework:

**i)** The **Mixture Model Network (Monet)**, as proposed by [36], is a broad spatial setting for multiple strategies specified on graphs or manifolds convolution operator in Monet is established for the given functions *g* & *f* defined on graph vertices as:

$$(f \star g) = \sum_{j=1}^{J} g_j D_j(v) f \quad (23)$$
$$D_j(v)f = \sum_{u \in \mathcal{F}_v} w_j(\mathbf{u}(v,u)) f(u) \quad (24)$$

The functions $w_1(u),...,w_J(u)$ apply weights to neighbours based on their pseudo-coordinates. As a result, *D<sub>j</sub>(v)f* is the aggregate value of the neighbour's functions. MoNet can instantiate numerous methods by defining distinct **u** & **w**. The functions *f, g* maps nodes to their attributes in GCN, whereas pseudo-coordinates of *u,v* are represented as $u(v,u) = (|N_v|,|N_u|)$, $J = 1$ & $w_1(u(v,u)) = \frac{1}{\sqrt{|N_y\| N_u|}}$ parameters *w<sub>j</sub>* are learnable in MoNet's model.

**ii) MPNN**: A generalised paradigm for spatial-based ConvGNNs is provided in **Message Passing Neural Networks (MPNN)** [37]. Graph convolution comprises of message-passing stage wherein data can be sent directly from one node to another along edges. To allow information to spread further, MPNN performs K-step message-passing iterations. So, Spectral Graph convolution employs learnable parameterised functions like message function M<sub>k</sub> to aggregate "message" *m<sup>k</sup><sub>v</sub>* from neighbours & update function U<sub>k</sub> to update hidden state *h<sup>t</sup><sub>v in</sub>* the message passing phase as:

$$\mathbf{m}_v^k = \sum_{u \in N(v)} M_k\left(\mathbf{h}_v^{(k-1)}, \mathbf{h}_u^{(k-1)}, \mathbf{x}_{vu}^e\right) \quad (25)$$
$$\mathbf{h}_v^{(k)} = U_k\left(\mathbf{h}_v^{(k-1)}, \sum_{u \in N(v)} M_k\left(\mathbf{h}_v^{(k-1)}, \mathbf{h}_u^{(k-1)}, \mathbf{x}_{vu}^e\right)\right) \quad (26)$$

Where $h_v^{(0)} = x_v$. In *the readout phase* of GCNN, hv<sup>(K)</sup> either be passed to an output layer or even to a readout function depending on whether one attempts to predict the node or graph level tasks respectively. After extracting hidden representations of each node. Readout function uses node-hidden representations to generate a representation of entire graph defined as:

$$\mathbf{h}_G = R\left(\mathbf{h}_v^{(K)} \mid v \in G\right) \quad (27)$$

Where **K** denotes the total number of time steps & *R(.)* denotes the readout function with learnable parameters. By presuming various forms of *U<sub>k</sub>(.), M<sub>k</sub>(.) & R(.)*, MPNN can cover several current GNNs, such as, [3], [38], [39]. However, earlier [40]. GIN modifies the central node's weight using a learnable parameter called ϵ<sup>(k)</sup> to compensate for this shortcoming. GIN performs graph convolutions using multi-layer perceptron *MLP as*:

$$\mathbf{h}_v^{(k)} = MLP\left((1+\epsilon^{(k)})\mathbf{h}_v^{(k-1)} + \sum_{u \in N(v)} \mathbf{h}_u^{(k-1)}\right) \quad (28)$$

**iii) Non-local Neural network (NLNN)** proposed by [18] combines multiple "self-attention"-style algorithms [41]

Table III. VARIOUS MODELS OF SPECTRAL CONVOLUTION GNNS

| Spectral Models | Definition | Math Representation | Advantages/Disadvantages |
|---|---|---|---|
| Spectral network [42] | The filter $g_\theta = \Theta_{i,j}^{(k)}$ is specified as a diagonal matrix learnable acquirable containing parameters. | Spectral Graph CNN is defined as: $$H_{:,j}^{(k)} = \sigma\left(\sum_{i=1}^{f_{k-1}} U\Theta_{i,j}^{(k)} U^T H_{:,i}^{(k-1)}\right)$$ $(j = 1, 2, \cdots, f_k)$ Wherein $k$ is layer index, $H_{:,i}^{(k-1)}$ is graph-signal input. Also, $H(0) = X$, $f_{k-1}$ & $f_k$ is the total count of input & output channels, respectively. | Eigen factorization of the laplacian matrix results in the following limitations: i) Any mutation to a graph causes a change in the eigen basis. ii) Learnt filters cannot be applied to a different graph topololgy. iii) computation-complexity is $O(n^3)$. |
| Chebyshev Spectral CNN (ChebNet) [43] | Filter $g_\theta$ is approximated via the Chebyshev polynomials of eigenvalues of a diagonal matrix. $g_\theta = \sum_{i=0}^{K} \theta_i T_i(\grave{\Lambda})$, where $\grave{\Lambda} = \frac{2\Lambda}{\lambda_{max}} - I_n$, values of $\grave{\Lambda}$ are in [-1 1]. | Chebyshev polynomials are defined recursively by $T_i(x) = 2xT_{i-1}(x) - T_{i-2}(x)$ with $T_0(x) = 1$ & $T_1(x) = x$. So, convolutional graph signal $x$ with defined filter $g_\theta$ is : $$x *_G g_\theta = U\left(\sum_{i=0}^{K} \theta_i T_i(\grave{\Lambda})\right) U^T x$$ $\grave{L} = \frac{2L}{\lambda_{max}} - I_n$, $T_i(\grave{L}) = UT_i(\grave{\Lambda})U^T$, which is provable by induction on $i$, ChebNet takes its ultimate form as: $$X *_G g_\theta = \sum_{i=0}^{K} \theta_i T_i(\grave{L})X$$ | i) computational complexity is $O(m)$ ii) ChebNet's filters can extract local features regardless of graph size, which is an improvement over Spectral CNN. |
| CayleyNet [44] | Cayley polynomials encompass a minimal frequency band as they are complex parametric rational functions. | CayleyNet's spectral graph convolution is defined as $$x *_G g_\theta = c_0 x + 2Re\left\{\sum_{j=1}^{r} c_j (hL - iI)^j (hL + iI)^{-j} x\right\}$$ Wherein $Re(.)$ results in real component for complex-number, $c_j$ & $c_0$ denotes complex & real coefficients respectively, $i$ denotes number is imaginary & $h$ depicts Cayley filter's spectrum-controlling factor. | ChebNet's ability to maintain spatial proximity demonstrates that it is a unique case of CayleyNet Graph. |
| Graph Convolution Network (GCN) [14] | Investigates first-order ChebNet equivalence. Assuming $\lambda_{max} = 2$ & $K = 1$. Ultimately ChebNet =n is reduced to: $x *_G g_\theta = \theta_0 x - \theta_1 D^{-\frac{1}{2}} A D^{-\frac{1}{2}} X$ | GCN further assume $= \theta_0 = -\theta_1$, resulting in following definition of graph convolution: $$x *_G g_\theta = \theta\left(I_n + D^{-\frac{1}{2}} A D^{-\frac{1}{2}}\right) x$$ For allowing multi-channel inputs/ outputs, GCN updates above =n into the compositional layer specified as: $H = X *_G g_\Theta = f(\grave{A} X \Theta)$ , Where $f(.)$ is activation function & $\grave{A} = I_n + D^{-\frac{1}{2}} A D^{-\frac{1}{2}}$ , this value of $\grave{A}$ causes instability in GCN. So, the normalisation process is used to replace the above $\grave{A}$ s value by $\hat{A} = \grave{D}^{-\frac{1}{2}} \grave{A} \grave{D}^{-\frac{1}{2}}$ with $\grave{A} = I_n + A$ & $\grave{D}_{ii} = \sum_j \grave{A}_{ij}$ | Limit number of parameters & overcome over-fitting GCN might even be viewed as a spatial-based aggregation of feature data from the neighbours of a node. The compositional graph Convolution finally can be written as $$h_v = f\left(\Theta^T\left(\sum_{u \in \{N(v) \cup v\}} \bar{A}_{v,u} x_u\right)\right) \forall v \in V$$ |
| Dual Graph Convolution Network (DGCN) [45] | DGCNN with two graph convolution layers running in parallel to capture local & global consistency & employs an unsupervised loss to ensemble them. | GCNN layers use normalised adjacency matrix $A$ & Positive Pointwise Mutual Information (PPMI) matrix, which captures node co-occurrence information via random walks taken from a graph. PPMI [45] is defined as: $$\mathbf{PPMI}_{v_1,v_2} = max\left(\log\left(\frac{count(v_1,v_2)\cdot|D|}{count(v_1)count(v_2)}\right),0\right)$$ where $v_1,v_2$ belongs to V, $|D| = \sum_{v_1,v_2} count(v_1,v_2)$ & *count(.)* function yields frequency at which vertices *u* and/or *v* co-exist in chosen random walks. | i) Encodes both local & global structural information without requirement to stack multiple graph convolution layers by combining outputs from dual graph convolution layers <br> ii) AGCN & DGCN strive to improve spectral approaches by augmenting Graph Laplacian. |
| GWNN [46] | graph wavelet neural network that replaces graph Fourier transform with graph wavelet transform without matrix decomposition | Spectral techniques are theoretically sound, & various theoretical studies have lately been proposed. These models can only be applied under "transductive" setting of graph tasks. | Benefits: <br> i) graph wavelets are easily created <br> ii) graph wavelets are sparse & localised, resulting in superior interpretable results [2]. <br> iii) outperforms on semi-supervised vertex classification procedure. |

[30],[31].In computer vision, NLNN generalises & extends non-local mean operation [47]. The latent state is calculated for a point as a weighted summation of attributes at all viable positions using non-local operations (space, time or both). Generalised non-local operation is defined in the same way as non-local mean operation:

$$\mathbf{h}_v^{t+1} = \frac{1}{\mathcal{C}(\mathbf{h}^t)}\sum_{V_u} f(\mathbf{h}_v^t, \mathbf{h}_u^t)g(\mathbf{h}_u^t) \quad (29)$$

Where $h^t_u$ calculates a scalar between *u* & *v* indicating relationship between them, $g(\mathbf{h}_u^t)$ indicates modification of input $\mathcal{C}(\mathbf{h}^t)$ & $h^t_u$ as a normalisation component, *u* represents the index over all potential positions for *v*. $f(\mathbf{h}_v^t, \mathbf{h}_u^t)$ represents a relation between *v* & *u* by computing a scalar between them. Different *f* & *g* choices can be used to define different NLNN versions.

**iv) Graph Network (GN):** GN offers a extensive framework employed for learning/acquiring of representations at edge, node & graph levels. Graph network (GN) [48] is a more general framework than others. It can unify MPNN, NLNN, CommNet [49], structure2vec [50], [51], GGNN, Interaction Networks [52], [53] Deep Sets [54], Point Net [55] Relation Network [56],[57] & so on. The GN block is main computation unit of GN, which specifies three different update *(φ)*, & aggregation *(ρ)* functions as:

$$\mathbf{e}_k^{t+1} = \varphi^e(\mathbf{e}_k^t, \mathbf{h}_{r_k}^t, \mathbf{h}_{s_k}^t, \mathbf{u}^t), \bar{\mathbf{e}}_v^{t+1} = \rho^{e \to h}(\mathbf{E}_v^{t+1}) \quad (30)$$
$$\mathbf{h}_v^{t+1} = \varphi^h(\bar{\mathbf{e}}_v^{t+1}, \mathbf{h}_v^t, \mathbf{u}^t), \bar{\mathbf{e}}^{t+1} = \rho^{e \to u}(\mathbf{E}^{t+1}) \quad (31)$$
$$\mathbf{u}^{t+1} = \varphi^u(\bar{\mathbf{e}}^{t+1}, \bar{\mathbf{h}}^{t+1}, \mathbf{u}^t), \bar{\mathbf{h}}^{t+1} = \rho^{h \to u}(\mathbf{H}^{t+1}) \quad (32)$$

All these functions can have a range of settings, must be invariant to input order & accept arguments of varying lengths. The sender vertex of edge *k* is $s_k$, while receiver node is $r_k$. At time step $t_{t+1}$, $E^{t+1}$ & $H^{t+1}$ denote matrices comprising stacked edge & node vectors respectively. With receiver node *v*, $E_v^{t+1}$ collects edge vectors & global graph representation attribute is *u*. Table IA *(where A represents the Appendix)* gives a detailed Overview of almost all ConvGNNs.

**2) Propagation modules - Recurrent Operator**

The main difference across the recurrent & the convolution operators is that convolution operator layers utilise separate weights whereas layers in [6] The equivalent weights of recurrent operators are shared. Prior approaches dependent on neural networks of recursive nature concentrate on handling directed acyclic graphs [58], [59] & are defacto standards in modelling sequential data. Thereafter the idea of a graph neural network (GNN) was introduced [60], that advanced established neural networks to process more graph types. RecGNN uses a repeating neural architecture to learn graph or node-level representations RecGNNs presume that nodes in the graph will constantly exchange information/messages with surrounding nodes until they establish a stable equilibrium & hence can capture both recursive & sequential graph patterns. Graph-level & node-level RNNs are the two main categories of graph RNNs. The primary difference is whether the motifs are discovered at the graph or node level for representing node states or a common graph state respectively. Table IV summarises the critical attributes of some approaches to RecGNN.

**a) Convergence-based Recurrent methods**

GNN aims to learn embedding-state $\mathbf{h}_v \in \mathbf{R}^s$ for every node that includes features about the neighbourhood &

itself. Convergence based Recurrent methods are usually used for learning Node-level patterns. The s-dimension vector of node *v* denotes embedding state **h**$_v$ utilised to generate output *o*$_v$ like the projected node label distribution. The HV & *o*$_v$ computation steps are therefore defined as follows:

$$\mathbf{h}_v = f(\mathbf{x}_v, \mathbf{x}_{co[v]}, \mathbf{h}_{\mathcal{N}_v}, \mathbf{x}_{\mathcal{N}_v}) \quad (33)$$
$$\mathbf{o}_v = g(\mathbf{h}_v, \mathbf{x}_v) \quad (34)$$

Wherein $\mathbf{x}_v, \mathbf{x}_{co[v]}, \mathbf{h}_{\mathcal{N}_v}, \mathbf{x}_{\mathcal{N}_v}$ are the essential characteristics of *v*, edge attributes, the states & parameters neighbouring nodes of *v*, respectively. Local transition function, *f*, denotes parametric function & modifies node states based on input from adjacent nodes and is common to all nodes. The local output function (G) specifies how the outcome is generated. Both of them are equivalent to feedforward neural networks. Consider **O, H, X**$_N$ **& X** are the matrices obtained through overlaying all outputs and states, & node features & graph-level attributes, respectively. Then there is a concise form as:

$$\mathbf{H} = F(\mathbf{H}, \mathbf{X}) \quad (35)$$
$$\mathbf{O} = G(\mathbf{H}, \mathbf{X}_N) \quad (36)$$

F & G symbolises the global transition & global output function, respectively. Assuming F is a contraction map, value H denotes the fixed point of the (31) and (32) & is determined uniquely. GNN computes the state using the following standard iterative approach, as suggested via fixed-point theorem [61] proposed by Banach, specified as $H^{t+1} = F(H^t, X)$ wherein $H^t$ represents *t*-th iteration for **H**. The dynamical system expressed by H$^{t+1}$ soon converges exponentially to the solutions for any initial value. Despite the fact that investigational results indicate that GNN is a suitable design for describing structural data, it nevertheless has significant drawbacks: The model's ability is limited as GNN demands *f* to serve as a contraction map. Also, it is ineffective to iteratively adjust the latent states of vertices towards their fixed point. Whenever the emphasis is on vertices instead of graphs, fixed points should be avoided since the representative distribution there at the fixed point will be precise in value & is impractical to use it to identify each node.

**i) GraphESN** [62] stands for Graph-Echo State Network, that generalises the **Echo State network** [63].It enhanced RecNN to handle cyclic/acyclic, undirected/directed but tagged graphs. It merely trains a reading function & employs a fixed contractive encoding function. The condensing of reservoir dynamics ensures convergence, ensuring GraphESN outperforms GNN in terms of efficiency.

**ii) Stochastic Steady-state Embedding (SSE)** [64] is another method for improving GNN efficiency. SSE presents a two-step learning system. In this, message propagation is facilitated via layer-wise feature extraction. During the update phase, a parametric operator modifies the embedding of each node, which are subsequently changed to the steady-state restriction space to fulfil the corresponding condition.

**iii) Lagrangian Propagation(LP-GNN)**[65] transforms the learning process into a Lagrangian local optimisation and does away with the need for fixed point recursive computations. A constraint satisfaction mechanism expresses the convergence procedure implicitly.

*b) Gate-based Recurrent methods*

Several research initiatives, including Long-Short Term Memory (LSTM) [66] and Gated Recurrent Units (GRU) [67], have sought to include a gate procedure in the propagation step to reduce the computation restrictions of GNN & increase long-term propagation of information across graph topology which is described below. Without any guarantee of convergence, they execute a fixed number of training steps. Gate-based methods are usually used for learning graph level patterns.

**i) Gated graph Neural Network (GGNN)** [68] removes the need for function *f* denotes contraction map cum recursive & instead propagates via GRU. It also computes gradients via Back Propagation Through Time (BPTT). GGNN computational stage is listed in Table V. Initially; vertex *v* aggregates adjacent messages. GRU-like update functions adjust the latent state of each node by utilising information from neighbouring nodes and the prior time step. Wherein $h_{Nv}$ represents collective information regarding the neighbourhood of node *v*, while r & z denotes reset & update gates

**ii) LSTMs** are deployed similarly to GRUs in the propagation phase, depending on a tree or graph.

**LSTM Tree:** The N-ary & the Child-sum Tree LSTMs are extensions of the basic LSTM architecture proposed by [69] or of already described recursive neural network-based models. Each vertex in Tree-LSTM involves data out of its children, constituting an instance of the graph. The Tree-LSTM elements of vertex v consist of forget gate *f*$_{vk}$ associated with every child k, rather than a total single forget gate as in normal LSTM. Table V illustrates the computing phase of Child-Sum Tree-LSTM. The output/input gates, & memory cell are represented by *o*$^t_v$, *i*$^t_v$ & *c*$^t_v$, respectively. The input vector at time step t is *x*$^t_v$. The Tree-LSTM (N-ary) is also developed for various trees in which each node has a maximum of *k*-ranked children. Table V. Introduces unique parameters for each child *k* in the equations for determining $h^{ti}_{Nv}$, $h^{tf}_{Nvk}$, $h^{to}_{Nv}$ & $h^{tu}_{Nv}$. Such parameter variables allow the model to acquire higher representations based on the behaviours of children of the module instead Child-Sum Tree-LSTM.

**iii) Graph LSTM**: Tree-LSTMs of both types are easily adaptable to the graph. Tree-LSTM (N-ary) implemented to the graph in [70] is an example of the graph-structured LSTM. However, since every vertex in the graph contains edges of indegrees not more than two. Peng et al. present a relation extraction task-based variation of the Graph LSTM. Because the edges of graphs in [71] have distinct labels, [71] use separate weight matrices to represent them. The edge label between nodes *v* & *k* is denoted by *m(v, k)* in Table V. To handle the semantic object parsing challenge, [68] propose a Graph LSTM network aids in choosing the beginning node & determines the node update sequence using a confidence-driven approach. It has a particular update sequence for generalising existing LSTMs to graph-structured datasets in, contrary to previously mentioned approaches.

**Sentence LSTM (S-LSTM)** is a text encoding technique proposed by [32]. It transforms text into a graph & learns the representation using the Graph LSTM.

**B. Skip Connection --Propagation Module**

Layers of neural graph networks are often combined in several applications to gain better results since more *(k)* layers lead each node to gather additional data from neighbours k hops away. In several experiments, it was discovered that deeper models did not increase performance as additional layers can propagate noisy data out of an exponentially growing amount of extended neighborhoods. All this leads to over-smoothing concerns since nodes seek similar representations after the aggregation procedure. To overcome this problem, "skip connections" to GNN models are added. Three different types of skip connections are discussed below:

**1) Highway GCN:** Scientists like [72] propose a Highway GCN that is analogous to highway networks in that it uses layer-wise gates [73]. With gating weights, a layer's output is added to its input as:

$$T(h^t) = \sigma(W_t h^t + b_t) \quad (37)$$
$$h^{t+1} = h^{t+1} \odot T(h^t) + h^t \odot (1 - T(h^t)) \quad (38)$$

In the case mentioned above, incorporating the highway gates improves performance at four layers[73]. The highway network is also used by the column network (CLN) [35]. However, the gating weights are computed using distinct functions.

**2) Jump Knowledge Network (JKN):** [74] investigate the features & drawbacks of neighbourhood aggregation techniques. JKN model is capable of learning adaptive & structure-aware representations. For each node in the last layer, JKN chooses among the whole set of intermediate representations, enabling the model to modify the size of the effective neighbourhood of each vertex as required. In their studies,[40] apply three ways to aggregate information: concatenation, max-pooling & LSTM-attention.JKN succeeds in bioinformatics, citation & social network investigations.& it can increase the performance of models like GCN, GraphSAGE & GAT.

**3) DeepGCNs:** ResNet [75] & DenseNet [76] are used as inspiration by [6] for discovering DeepGCNs. ResGCN & DenseGCN are developed for overcoming the problems of vanishing gradient & over smoothing, which incorporates residual connections & dense connections. The hidden node state in both can be expressed as:

$$h_{Res}^{t+1} = h^{t+1} + h^t \quad (39)$$
$$h_{Dense}^{t+1} = \|_{i=0}^{t+1} h^i \quad (40)$$

DeepGCNs experiments are done on the semantic segmentation of point clouds, & its performance is improved with a 56-layer model.[77], for instance, added the following residual connections to Kipf & Welling's first-order GCN filter equation. Through experiments, they demonstrated that incorporating such residual connections could enable a deeper network, which is comparable to ResNet's findings.

$$h_i^{l+1} = \alpha_i^l \odot \tilde{h}_i^{l+1} + (1 - \alpha_i^l) \odot h_i^l \quad (41)$$

Where $\alpha_i^l$ a set of weights is determined as follows and $h_i^{l+1}$ is calculated according to Eq. (14). Other variants used as skip connection propagation modules are:

**Column Network (CLN) [35]:** It uses a similar approach & incorporates the below residual connections using adjustable weights:

$$\alpha_i^l = \rho\left(b_\alpha^l + \Theta_\alpha^l h_i^l + \Theta_\alpha' \sum_{j \in \mathcal{N}(i)} h_j^l\right) \quad (42)$$

Where $b_\alpha^l$, $\Theta_\alpha^l$ & $\Theta_\alpha'$ are parameter variables. Observe that (37) is quite analogous to GRU found in GGS-NNs [78]. Variations are that the superscripts in a CLN indicate the count of layers, and various layers have varying parameter values. However, in GGSNNs, pseudo-time is represented as the superscript, and only a group of variables are utilised during numerous time steps. In a heterogeneous graph, CLN averaged the findings after training various parameters for various edge-type discrete features.

**PPNP**[76]:Designed graph convolutions with teleportation to the first layer as inspired by customised PageRank as:

$$H^{l+1} = (1 - \alpha)\tilde{D}^{-\frac{1}{2}}\tilde{A}\tilde{D}^{-\frac{1}{2}}H^l + \alpha H^0 \quad (43)$$

Where $\alpha$ is a hyper-parameter & $H_0 = \mathcal{F}_\theta(F^V)$ & $\mathcal{F}_\theta(.)$ contains all the parameters as opposed to the graph convolutions.

**C. Sampling Modules**

The GNN model gathers messages from neighbours of nodes in the previous layer. Going back a few GNN layers, it appears that the size of the supported neighbours expands exponentially ("neighbours exploding") with depth. It is also impossible to gather messages like this while working with massive graphs. Sampling is a cost-effective & efficient method to address the same & is used in propagation to ensure that all nearby information for each node is always kept & processed. The different types of graph sampling modules described in this section, as shown in *Fig. 9.* are as under:

**1) Node sampling**

The simple strategy to minimise the extent of neighbourhood nodes is designating a subgroup of each node's neighbours. GraphSAGE [79] samples a limited number of neighbours for each node, maintaining a neighbourhood size of 2 to 50. This method leverages the previous hidden state as an inexpensive approximation to restrict the receptive field in the neighbourhood of 1-hop.As presented in [80], the importance-based sampling technique identifies $T$ nodes associated with the greatest normalised visit counts via generating random walks from the target nodes.

**2) Layer sampling**

To limit the scale of the expansion, layer sampling maintains a fixed count of nodes in each layer from which to get aggregate data. FastGCN [45] samples each layer's receptive field directly & employs importance sampling most (only significant nodes are sampled from each layer).

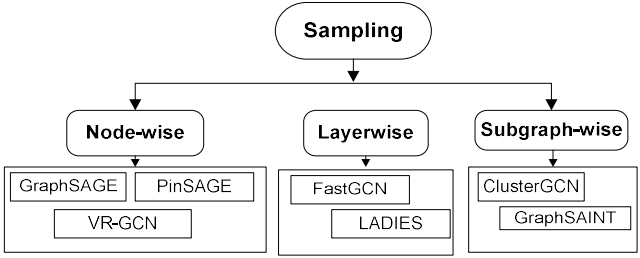

**Fig. 9.** Classification of Sampling Module

Huang et al. (2018) provide a parametric and trainable sampler for a layer-by-layer sampling based on the preceding layer. This sampler [81] optimises sampling importance while reducing variation at the same time. LADIES [82] layer-wise sampling sparsity problem via designing samples from union of neighbours of a node.

### 3) Subgraph sampling

This approach sample numerous subgraphs & limit the neighbourhood search to these subgraphs from the entire graph. Various models like ClusterGCN [83] and GraphSAINT [84] have been proposed that work on the principle of graph clustering techniques to sample subgraphs and samples nodes or edges directly to build a subgraph, respectively.

### D. Graph Pooling Modules

The CNN layer is frequently followed by a pooling layer in computer vision that provides more generic features. A rich hierarchy is common in complex & large graphs that are critical for classification tasks at graph levels as node features need to be aggregated for useful information than at the node level focused tasks. Such procedures are usually known as Readout or Pooling operations.

Traditional CNN's employ several stride convolutions or pooling to decrease resolution. Often the work is centered on building hierarchical pooling layers on the graph, just like these pooling layers. The two types of pooling modules discussed in this section are direct pooling modules & hierarchical pooling modules, as shown in Fig.10.

### 1) Direct Pooling Modules

With different node selection algorithms, the Direct Pooling Module learns refined graph-level structures explicitly from nodes. Such units in certain versions are referred to as reading functions

**a) Simple node pooling:** These models use per-node max/average/sum/attention operations to obtain a global network representation of the node features.

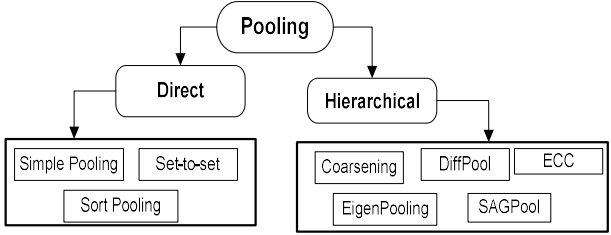

**Fig. 10.** Classification of Pooling Modules

**b) Set2set:** MPNN employs the Set2set algorithm [85] as a read function to obtain the graph representation, Set2set is a tool that employs LSTM- eigen decomposition's inefficiency.

**c) SortPooling:** SortPooling [86] sorts node embedding according to their structural functions & then uses CNNs to generate the representation from the sorted embedding.

### 2) Hierarchical Pooling Unit/Module

Graphs exhibit complex hierarchical patterns in addition to having node & graph-level structures [54], which may be investigated using hierarchical clustering methods. For instance, [42] employed a density-based agglomerative clustering, while [87] used a multi-resolution spectral clustering [88]. The following are the methods based on hierarchical pooling:

**a) Graph Coarsening:** Most early techniques are based on graph coarsening algorithms. The first step is to use spectral clustering algorithms, which are inefficient due to the Eigen decomposition step. Graclus [89] is a pooling module that offers a faster approach for clustering nodes. ChebNet & Monet, for example, utilise graclus to combine pairs of nodes and add extra nodes to produce a balanced binary tree from pooling.

**b) ECC:** The pooling module of Edge-Conditioned Convolution (ECC) [90] is designed with a recursively down-sampling procedure. The down-sampling approach separates the graph into sub-groups depending on the sign of the greatest Laplacian eigenvector. Additionally, trained on edge type-based parameters, ECC used them to classify graphs.

**c) DiffPool:** DiffPool [91] employs a learnable hierarchical clustering module that involves each layer training an assignment matrix $S^t$

$$\mathbf{S}^t = \text{softmax}\left(\text{GNN}_{t_{\text{pool}}}(\mathbf{A}^t, \mathbf{H}^t)\right) \qquad (44)$$

$$\mathbf{A}^{t+1} = (\mathbf{S}^t)^T \mathbf{A}^t \mathbf{S}^t \qquad (45)$$

$H^t$ denotes vertex attribute matrix; $S^t$ depicts the probability that layer $t$ vertex is allotted to the coarser vertex in the layer $t+1$ & $A^t$ denotes adjacency-matrix in coarsened form for layer $t$

**d) gPool:** gPool [92] learns scores of projection of each vertex & selects top-k score-bearing nodes using a project vector. It employs a vector instead of a matrix at each tier, which minimises the storage complexity compared to DiffPool. However, the graph structure is ignored by the projection technique.

**e) EigenPooling:** EigenPooling [17] is a method for combining node characteristics with local structure. It extracts subgraph information via the local graph Fourier transform & suffers from graph eigen decomposition's inefficiency.

**f) SAGPool:** SAGPool [93] is another method for learning graph representations that combine features & topology. It employs an approach based on self-attention with an appropriate time & space complexity.

## IV. OTHER GRAPH NEURAL NETWORK MODELS - GRAPH AUTO-ENNCODERS

**Graph Auto Encoder (GAE)** specifies the architecture for deep neural networks using unsupervised learning, which encodes graphs, and nodes into hidden vector spaces & regenerates graph data out of hidden information. GAE is used to learn network embedding & generated distribution patterns through graph Network Embedding & Graph Generation processes, respectively, as described below. Table V summarises the main properties of some GAEs. Graphs are presumed to possess a potential low-rank nonlinear structure. In-depth descriptions of graph auto encoders, graph variational autoencoders and other innovations are provided in this section.

Network embedding describes node representation in a lower dimensional vector that retains topological information of the node. As illustrated in Figure 11, GAEs employ get network embeddings (1-hot Encoding) and a decoder to ensure network embeddings maintain graph topological information like PPMI & adjacency matrix. Table V below summarises the main characteristics of various models Graph Auto Encoder Models. Previously, GAEs for network embedding learning mainly was built using multi-layer perceptrons. In sparse autoencoder (SAE), AEs for graphs first emerged [94] by employing an adjacency matrix as well as its variants as crude node attributes & employed the L2-reconstruction error function as:

$$\min_{\Theta} \mathcal{L}_2 = \sum_{i=1}^{N} \| \mathbf{P}(i,:) - \hat{\mathbf{P}}(i,:) \|_2 \quad (46)$$

$$\hat{\mathbf{P}}(i,:) = \mathcal{G}(\mathbf{h}_i), \mathbf{h}_i = \mathcal{F}(\mathbf{P}(i,i)) \quad (47)$$

Wherein $\mathbf{h}_i \in \mathbb{R}^d$ denotes low-dimension vertex $v_i$ representation, $d \ll N$ denotes dimensionality,` P symbolises transition matrix, and $\Theta$ denotes parameter variables. F() and G() denote encoder & decoder, respectively. The MLP used for the encoder and decoder has numerous hidden layers, i.e. SAE reduces the data in P(i;:) to a low-dimension vector $\mathbf{h}_i$, and then utilises that to rebuild the original feature. An additional regularisation term for sparsity and k-means is employed for node clustering. The results show that SAEs perform better than non-deep learning baselines. However, SAE effectiveness's underlying mechanism is still unknown. Deep Neural Network for Graph Representations (DNGR) [95] encodes & / or decodes a PPMI matrix employing MLP and denoising stacked autoencoder [96]. In addition, Structural Deep Network Embedding (SDNE) [97] use a stacked-autoencoder for preserving first cum second-order closeness of vertices. By minimising the distance between network embeddings of a node & its neighbours, the first error function in SDNE for encoder output allows the learnt network embeddings to retain first-order proximity of a node (local pair-wise resemblance across nodes linked by edges).$L_{1st}$, the first loss/error function is specified as follows:

$$L_{1st} = \sum_{(v,u)\in E} A_{v,u} \|enc(\mathbf{x}_v) - enc(\mathbf{x}_u)\|^2 \quad (48)$$

wherein *enc(.)* denotes encoder consisting of the MLP & $x_v = A_v$. By demonstrating that the L2-reconstruction error in (45) relates to the 2nd-order proximity across nodes (that is, if two vertices share similar neighbours, they have identical latent representations). Collaborative filtering and structural deep network embedding (SDNE) [98] are network research concepts that solve the paradox.

.A new Laplacian eigenmap term was added to the goal function by SDNE in response to network embedding's findings that the first-order closeness is also significant [97]. So, identical latent representations indicate a direct connection between two nodes. The second error function enables learnt network embedding to minimise the difference across inputs to a node & its reconstructed inputs, hence preserving the node's 2nd-order proximity. The adjacency matrix alters L2-reconstruction loss by employing weights for zero and nonzero elements. The second error function, $L_{2nd}$, is expressed as:

$$L_{2nd} = \sum_{v\in V} \|(dec(enc(\mathbf{x}_v)) - \mathbf{x}_v) \odot \mathbf{b}_v\|^2 \quad (49)$$

Wherein *dec(.)* denotes the decoder consisting of an MLP & $b_{v,u} = 1$ if $A_{v,u} = 0$, $b_{v,u} = \beta > 1$ if $A_{v,u} = 1$.

The input matrix construction in SDNE has a time complexity of $O(N^2)$, hence not scalable to large-scale graphs. SDNE [99] & DNGR [100] consider Only node structure information, such as the connectivity between pairs of nodes. They neglect node feature information that shows node properties. Graph Autoencoder **(GC-MC)** uses GCN [101] to encode node structure & feature information simultaneously. The **GC-MC** encoder is made up of two graph convolution layers as

$$\mathbf{Z} = enc(\mathbf{X}, \mathbf{A}) = Gconv(f(Gconv(\mathbf{A}, \mathbf{X}; \Theta_1)); \Theta_2) \quad (50)$$

Whereas **Z** denotes the embedding matrix of graph network, *Gconv(.)* & *f(.)* functions denote the graph convolution layer & ReLU activation function, respectively. The goal of GAE's decoder is to recover information about nodes via underlying embeddings, described as:

$$\hat{\mathbf{A}}_{v,u} = dec(\mathbf{z}_v, \mathbf{z}_u) = \sigma(\mathbf{z}_v^T \mathbf{z}_u) \quad (51)$$

Where $z_v$ is node $v$'s embedding, a GAE trained via reducing negative-cross entropy while considering adjacency-matrix, *A* and rebuilt adjacency-matrix, $\hat{\mathbf{A}}_{v,u}$.

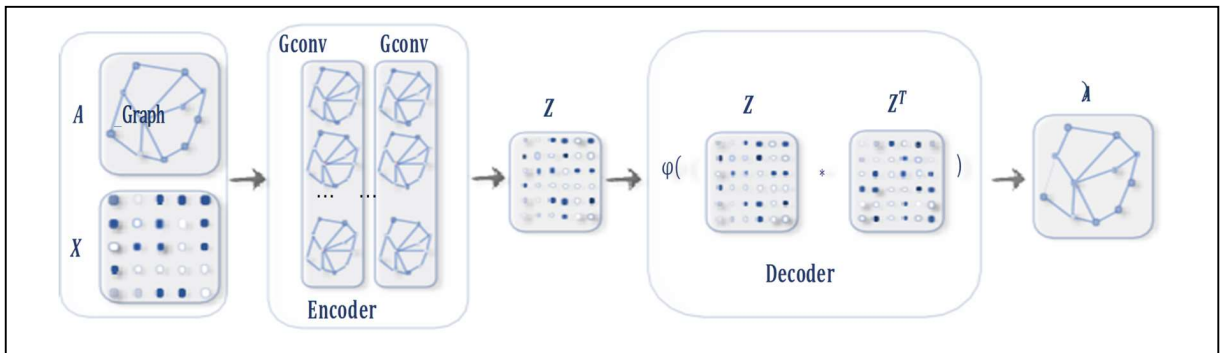

**Fig. 11.** A GAE for network embedding [101]

TABLE V: PRIMARY FEATURES OF VARIOUS GAEs

| Approaches | Target-Objective | Input(s) | Encoder | Decoder |
|---|---|---|---|---|
| *DNGR* [102] | PPMI –matrix reconstruction | A | MLP | MLP |
| *SDNE* [35] | Retain first cum second-order proximity of nodes. | A | MLP | MLP |
| *GAE\** [4] | Adjacency-matrix reconstruction | A, X | ConvGNN | Similarity-metric |
| *VGAE* [4] | Learning the generative distribution of data | A, X | ConvGNN | Similarity-metric |
| *ARVGA* [94] | Learning the generative distribution of data adversarially | A, X | ConvGNN | Similarity-metric |
| *DNRE* [90] | Retrieve network-embedding | A | Graph-LSTM | Identity-function |
| *NetRA* [28] | Retrieve network embedding via adversarial training | A | Graph-LSTM | LSTM-network |
| *DeepGMG* [103] | Maximise the joint expected log-likelihood | X, A, $X^e$ | RecGNN | Decision procedure |
| *GraphRNN* [103] | Maximise the permutation-likelihoods | A | RNN | Decision procedure |
| *GraphVAE* [104] | Minimise the reconstruction-loss | X, A, $X^e$ | ConvGNN | MLP |
| *RGVAE* [105] | Minimise the reconstruction loss via validity constraints | X, A, $X^e$ | CNN | Deconvolution net |
| *MolGAN* [106] | Optimise generative adversarial & RL-losses | X, A, $X^e$ | ConvGNN | MLP |
| *NetGAN* [107] | Optimise generative adversarial-loss | A | Graph-LSTM | Graph-LSTM |

Rebuilding the adjacency matrix for graphs potentially results in overfitting due to the ability of autoencoders. Another line of research addresses data sparsity while learning network embedding by transforming the graph into sequences using random walks or even its permutations.
**Deep Recursive Network Embedding (DRNE)**[108] proposes that vertex network embeddings must be very similar to aggregation of embedding of its neighbourhood done through a Long Short-Term Memory (LSTM) network [66]. The DRNE reconstruction error is defined as:

$$L = \sum_{v \in V} \| z_v - \text{LSTM}(\{z_u \mid u \in N(v)\}) \|^2 \quad (52)$$

The LSTM Graph network accepts an arbitrary sequence of neighbours of vertex *v* ordered on respective node degrees as inputs, using a local neighbour-sampling approach for vertices of high degrees to eliminate unnecessarily large memory. Also, $z_v$ specifies network embedding for vertex *v* generated via a look-up of the dictionary. DRNE automatically acquires network embeddings through the LSTM network instead of employing the graph-LSTM to derive network embeddings, as stated in (48). It overcomes the issue of graph-LSTM not invariant to node sequences' permutation. NetRA offers a graph encoder-decoder architecture with the following generalised equation:

$$L = -\mathbb{E}_{z \sim P_{data}(z)}(\text{dist}(z, \text{dec}(\text{enc}(z)))) \quad (53)$$

Where the difference between the vertex embedding *z* & its reconstructed *z* is quantified ***dist(.)***, NetRA's encoder & decoder use LSTM networks as inputs, containing random walks anchored upon every vertex *v* ∈ *V*. NetRA, like ARVGA [109], uses adversarial training to regularise acquired network embeddings within the preceding distribution. The experimental findings validate NetRA's effectiveness, although it skips the node permutation variation issue of LSTM networks.

**Graph2Gauss(G2G)**[110] recommended encoding every vertex as a Gaussian distribution, $h_i = \mathcal{N}(M(i,:), \text{diag}(\Sigma(i,:)))$ which highlights the uncertainty of vertices as opposed to the above research that translates vertices to a low-dimensional vector. The researchers adopted the following deep mapping as the encoder, relating the vertex attributes to variances & means of the Gaussian distribution:

$$M(i,:) = \mathcal{F}_M(F^V(i,:)), \Sigma(i,:) = \mathcal{F}_\Sigma(F^V(i,:)) \quad (54)$$

The parameters that need to be learnt are $\mathcal{F}_M()$ and $\mathcal{F}_\Sigma()$. The model was then learnt using pair-wise constraints rather than an explicit decoder function:

$$\text{KL}(h_j \| h_i) < KL(h_{j'} | h_i)$$
$$\forall i, \forall j, \forall j' \text{ s.t. } d(i,j') > d(i,j) \quad (55)$$

Wherein $d(i, j')$ symbolises shortest-distance between vertices $v_i$ and $v_j$. Also, ***KL(q(.) ∥ p())*** represents divergence across q() and p(). The restrictions guarantee that the graph distance & the KL-divergence across vertex representation have the equivalent relative order. Equation (51) is challenging to optimise, hence a relaxation using an energy-based loss [111] was used instead:

$$\mathcal{L} = \sum_{(i,j,j') \in \mathcal{D}} \left( E_{ij}^2 + \exp^{-E_{ij'}} \right) \quad (56)$$

$\mathcal{D} = \{(i, j, j') \mid d(i, j') > d(i, j)\}$ & $E_{ij} = \text{KL}(h_j \| h_i)$, a strategy of unbiased sampling is used to accelerate the training procedure.

**Variational Autoencoders**
Variation autoencoders (VAEs) differ from earlier autoencoders as they incorporate dimensionality reduction via generative models. Tolerating noise and learning smooth representations are two of its possible advantages [16].VGAE [16], where the decoder represents the primary linear product, was the first to use VAEs with graph data. The variational variant of GAE, *i.e.*VGAE [101], can be used to learn the data distribution. The variational lower limit $L$ is optimised via VGAE:

$$L = E_{q(Z|X,A)}[\log p(A \mid Z)] - KL[q(Z \mid X, A) \parallel p(Z)] \quad (57)$$

$KL(.)$ measures the distance between two distributions & is known as the Kullback-Leibler divergence function, Gaussian prior $p(Z) = \prod_{i=1}^{n} p(z_i) = \prod_{i=1}^{n} N(z_i \mid 0, I)$, $p(A_{ij} = 1 \mid z_i, z_j) = \text{dec}(z_i, z_j) = \sigma(z_i^T z_j)$, $q(Z \mid X, A) = \prod_{i=1}^{n} q(z_i \mid X, A)$ with $q(z_i \mid X, A) = N(z_i \mid \mu_i, \text{diag}(\sigma_i^2))$.

Wherein the mean vector ($\mu_i$) denotes *i*th-row for encoder results as described by(46) & log $\sigma_i$ is computed the same way as $\mu_i$ with another encoder. VGAE supposes that the real-distribution $q(Z \mid X, A)$ ought to be near the preceding distribution $p(Z)$. However, this method's temporal complexity is $O(N^2)$ because it requires recreating the entire graph. DVNE [98] offered a different VAE employed on graph data representing each vertex as a Gaussian distribution in nature, which SDNE and G2G inspire. To ensure the transitivity of similarities for nodes, DVNE employed Wasserstein distance instead of other efforts that employed KL-divergence for metric. DVNE's objective function also retained first cum second-order proximity denoted as:

$$\min_{\Theta} \sum_{(i,j,j') \in \mathcal{D}} \left( E_{ij}^2 + \exp^{-E_{ij'}} \right) + \alpha \mathcal{L}_2 \quad (58)$$

where $E_{ij} = W_2(h_j \parallel h_i)$ denotes the second Wasserstein distance across two Gaussian distributions $h_j$ & $h_i$. $\mathcal{D} = \{(i,j,j') \mid j \in \mathcal{N}(i), j' \notin \mathcal{N}(i)\}$ denotes a combination of triples connected to Ist-order proximity ranking error. Following is a description of the reconstruction error:

$$\mathcal{L}_2 = \inf_{q(Z|P)} \mathbb{E}_{p(P)} \mathbb{E}_{q(Z|P)} \parallel P \odot (P - \mathcal{G}(Z)) \parallel_2^2 \quad (59)$$

Where Z stands for samples taken from H, and P is the transition matrix. By employing the reparameterisation strategy, the objective function can be minimised using this method, just like in traditional VAEs [28].

**Improvements and Discussions**

In ARGA [112], an additional regularisation term known as adversarial training strategy is included in GAEs. In particular, the discriminator sought to determine if a latent depiction originates via a generator or with a prior distribution, whereas the of GAEs served as the generator. Therefore, as regularisation, the autoencoder must conform to the previous distribution. The objective function is:

$$\min_{\Theta} \mathcal{L}_2 + \alpha \mathcal{L}_{GAN} \quad (60)$$

$$\text{Where} \mathcal{L}_{GAN} = \min_{G} \max_{D} \mathbb{E}_{h \sim p_h}[\log \mathcal{D}(h)] + \mathbb{E}_{z \sim G(F^V, A)}[\log(1 - \mathcal{D}(z))]$$

& $L_2$ is reconstruction loss in GAE. Using the graph convolution encoder from (53), where the generator is specified by $G(F^V, A)$, $p_h$ is the prior distribution & $\mathcal{D}()$ denotes the discriminator dependent on cross entropy error. The experiment results showed that the adversarial training method was effective, and the study used a straightforward Gaussian prior.

NetRA [113] & the generative adversarial network (GAN) also suggests graph autoencoders' potential to generalise. The authors utilised the following objective function in particular:

$$\min_{\Theta} \mathcal{L}_2 + \alpha_1 \mathcal{L}_{LE} + \alpha_2 \mathcal{L}_{GAN} \quad (61)$$

Where $\mathcal{L}_{LE}$ is the objective function for the Laplacian eigenmaps described in (54), the authors also used an LSTM as the encoder to combine data from areas that are comparable to (48). The scientists employed random walks to construct the sequences of input rather than sampling only close neighbours & arranging the vertices by degrees like in DRNE [114].In contrast to ARGA, NetRA employed Gaussian noises followed via MLP acting as a generator and considered the depictions in GAEs as the actual truth.

If vertex attributes are encoded into the encoder, GAEs can be used in the inductive learning environment similarly to GCNs. This can be accomplished in two ways: either by explicitly learning a function of mapping from node attributes, like in G2G [115] or by employing an encoder of GCN as in VGAE [116], GC-MC [16] and VGAE [117]. Even if a node is not spotted during the training phase, the model can still be utilised as the information of edges is utilised only while learning parameter variables. These studies also demonstrate that, although GAEs & GCNs rely on various designs, it is feasible to deploy them in combination. Numerous similarity metrics have been used in GAEs, such as the Wasserstein distance and KL divergence for graph VAEs & ranking loss, L2-reconstruction loss, and laplacian eigenmaps for graph AEs. The best similarity metric for a specific task and model architecture still needs to be discovered, although these measures are based on many motivations.

### V. OTHER GRAPH NEURAL NETWORK-- SPATIAL-TEMPORAL NETWORKS

**Spatio-Temporal Graph Neural Networks** (STGNNs) contribute significantly to catching graph dynamics, including both structure & inputs of the graph, using a combination of convolution & recurrent operations. Modelling the dynamic inputs of the node while ensuring interdependency between interconnected nodes is the goal of these methods for capturing a graph's spatial & temporal dependencies at the same time. STGNNs are utilised for forecasting future node or spatial-temporal graph tags or values. STGNNs are differentiated into two types, i.e. RNN & CNN-based approaches. RNN-based techniques mostly incorporate spatial-temporal relations by employing graph convolutions to filter input & latent states fed to a recurrent module [32], [118], [20]. To demonstrate, consider the following Recurrent neural network:

$$H^{(t)} = \sigma(WX^{(t)} + UH^{(t-1)} + b) \quad (62)$$

wherein $X^{(t)} \in R^{n \times d}$ is vertex/node attribute-matrix at time-step $t$. Following the inclusion of graph-convolution, the above equation results in the following:

$$H^{(t)} = \sigma(Gconv(X^{(t)}, A; W) + Gconv(H^{(t-1)}, A; U) + b) \quad (63)$$

Wherein Gconv( ) denotes graph convolution layer. An LSTM network is combined with ChebNet [43] in the Graph Convolutional Recurrent Network (GCRN)

[118].Diffusion Convolutional Recurrent Neural Network (DCRNN) [20] employs unique convolution based on DGC output and encoder-decoder layout for predicting the very next *K*-steps for values of nodes. The recurrent framework proposed by Structural-RNN [119] predicts vertex tags at every time step. Each node's/ edge's temporal information is transferred through the respective node/ edge -RNN. Edges or nodes share the RNN model with the equivalent semantic

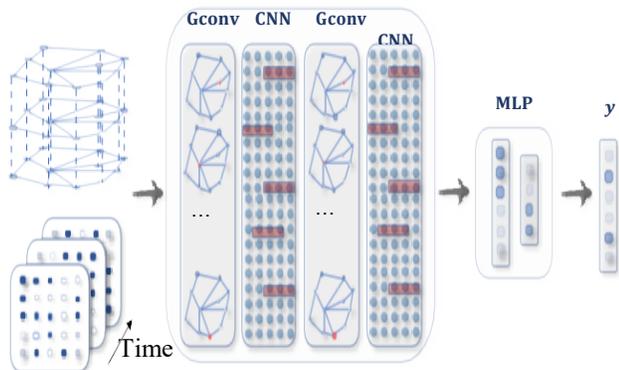

Fig. 12. Spatial-Temporal GNN [114]

Convolutional Recurrent Neural Network (DCRNN) [20] employs unique convolution based on DGC output and encoder-decoder layout for predicting the very next *K*-steps for values of nodes. The recurrent framework proposed by Structural-RNN [119] predicts vertex tags at every time step. Each node's / edge's temporal information is transferred through the respective node/ edge -RNN. Edges or nodes share the RNN model with the equivalent semantic class to reduce complexity. Node-RNN uses outcomes of edge-RNNs as input data to include spatial information. Time-consuming iterative propagation & gradient explosion/vanishing difficulties plague RNN-based techniques.

With the advantages of parallel processing, sustainable gradients & minimal memory demands, algorithms relying on CNN can accommodate spatiotemporal graphs through a non-recursive approach. CNN-based approaches intermingle1-dimensional CNN layers by graph convolution layers to learn temporal & spatial relationships, as shown in **Fig. 12**. Suppose the input signals to a spatio-temporal GNN are denoted by tensor $X \in R^{T*n*d}$; comparatively, the 1-dimensional layer of CNN drifts across $X_{[:, i, :]}$ along the time-axis to collect temporal data for each vertex, whereas the layer of graph convolution acts on $X_{[i, :, :]}$ for structural-data aggregation during every time step. CGCN [114] mixes 1D convolutional layers with layers from GCN [14] or ChebNet.It constructs a spatio-temporal module by sequentially mounting two gated 1D-convolution layers between a graph convolution layer. ST-GCN [24] uses a *1D convolutional layer & a PGC layer* to create a spatial-temporal block.

**Graph WaveNet** [120] presents the self-adaptive adjacency matrix to execute graph convolutions to learn latent static graph topologies from multiple snapshots of graph datasets together in a spatio-temporal scenario contrary to previous methods for using pre-defined graph structure to reflect the actual dependencies between nodes. The self-adaptive adjacency matrix has the following definition:

$$\mathbf{A}_{adp} = \text{SoftMax}(\text{ReLU}(\mathbf{E}_1 \mathbf{E}_2^T)) \quad (64)$$

Where $\mathbf{E}_1$ & $\mathbf{E}_2$ represent the source & target vertex embedding, respectively, using learnable parameters & the computation of the soft-max function along the row dimension. The dependency weight between a source & a target node can be calculated by multiplying *E1* by *E2*. Graph WaveNet achieves good results without an adjacency matrix via a sophisticated spatiotemporal neural network based on CNN. Researchers can use latent static spatial dependencies to find interpretable & durable correlations between distinct components in a network. On the other hand, learning latent dynamic spatial dependencies may increase model precision. g. in a transportation network, current traffic conditions can affect travel time between two roadways.

## VI. OTHER GRAPH NEURAL NETWORKS-- GRAPH REINFORCEMENT LEARNING

Reinforcement learning (RL) is the deep learning class recognised for its effectiveness in learning from feedback signals, especially when confronted with non-differentiable targets & strict limits like AI activities, e.g. playing video games [121]. Goal-directed molecular graphs were produced by GCPN [122] using RL, while non-differential goals &constraints were considered. Adding nodes & edges to a graph is designed as the **Generative model;** in particular, & **the Markov decision process** functions as RL agents in the graph generation setting. End-to-end learning of GCPN is possible using a policy gradient by examining agent actions for predictions of edge/ connection existence. It also employs adversarial & domain-specific rewards & for vertex representations, GCNs are used [123]. Table VII below provides the main features of various models of Graph Reinforcement Learning. A related study, **MolGAN** [124], also used RL to produce molecular graphs. However, it directly constructs the entire graph instead of a series of steps & is effective for small molecules. To forecast byproducts of chemical reactions, **GTPN** [125]adopted RL, where the agent specifically predicted new bonding types of selected node pairs in the molecular graph & is rewarded both immediately & at the conclusion based on the accuracy of the predictions. GTPN can employ a GCN to acquire node descriptions, whereas an RNN can be used to recall prediction sequences. **GAM** [126] utilised random walks to apply RL for graph categorisation. The random walk production is regarded as a Partially Observable Markov Decision Process (POMDP). An agent makes a prediction of tag for the node & then it chooses the upcoming node within a random walk. The ability of the agent to classify the graph accurately is expressed as:

$$\mathcal{J}(\theta) = \mathbb{E}_{P(S_{1:T};\theta)} \sum_{t=1}^{T} r_t \quad (65)$$

Wherein $r_t = 1$ denotes a correct prediction; else, $r_t$ = -1. T & $S_T$ indicates the total count of time steps & the environment, respectively.

**RL** was used for knowledge graph (KG) reasoning by

**DeepPath** [127] & **MINERVA** [54]. **DeepPath** focuses on determining the most valuable route between two target vertices, while MINERVA collaborated on question-answering challenges. Each step in both approaches requires RL agents to forecast the following node in the path & produce a line of reasoning to KG. If paths lead to intended locations, agents are rewarded. In order to support path variability, DeepPath additionally incorporated a regularization term.

## VII. OTHER GRAPH NEURAL NETWORK-- GRAPH ADVERSARIAL METHODS

In recent times, the researchers have focused more on adversarial training approaches, such as GANs and adversarial assaults. Graph Adversarial Methods are studied under the two categories as depicted below. Table VIII provides a summary of the key characteristics of graph adversarial methods.

### A. Adversarial Training

Building a discriminator & a generator as two linked models is fundamental concept behind a GAN. Discriminator seeks to determine whether a sample is developed by generator or it is derived from original data, while generator seeks to "hoodwink" discriminator by creating fake data. The result is that by employing a min-max game for joint training, both models gain from one another. It has been demonstrated that adversarial training improves generalization capacity of discriminative models &is efficient in generative models. Here, we go through a number of alternative adversarial training techniques for graphs in further depth. With following objective function, GraphGAN [128] suggested employing a GAN to improve graph embedding techniques [129].

$$\min_{\mathcal{G}} \max_{\mathcal{D}} \sum_{i=1}^{N} \left( \mathbb{E}_{v \sim p_{graph}(\cdot|v_i)}[\log \mathcal{D}(v, v_i)] + \mathbb{E}_{v \sim \mathcal{G}(\cdot|v_i)}[\log(1 - \mathcal{D}(v, v_i))] \right) \quad (66)$$

The generator $G(.)$ & discriminators $D(.)$ are specified as:

$$\mathcal{D}(v, v_i) = \sigma(\mathbf{d}_v \mathbf{d}_{v_i}^T), \mathcal{G}(v \mid v_i) = \frac{\exp(g_v g_{v_x}^T)}{\sum_{v' \neq v_i} \exp(g_{v'} g_{v_1}^T)} \quad (67)$$

where $d_v$ & $g_v$ denote low-dimension embedding vectors of vertex $v$ in discriminator & the generator, respectively. Discriminator makes it possible that node pairs in original graph should have high similarity. While node pairs built via generator must have minimal similarity when the (62) & (63) are combined. Only difference between this design & network embedding techniques like LINE [130] is that generator $G(.)$ builds negative node pairs rather than by random sampling & it improves node embedding vector capacity for prediction.

In order to enhance network-embedding techniques, adversarial network embedding (ANE) [130] also adopted an adversary training technique via implementing a prior distribution as actual data & considering embedding-vectors as the samples generated. It employs GAN as an extra component to already existing network embedding methods like DeepWalk. A GAN was utilized by **GraphSGAN** [131] to improve graph-based semi-supervised learning. To lessen propagation effect of current models across various clusters, bogus vertices should be designed in subgraph density gaps. A unique optimization goal is established with complicated error factors to guarantee that generator produces specimens in density gaps at equilibrium.GAN framework was used by **NetGAN** [115] to create & differentiate between random walks using an LSTM for graph generating tasks by learning distribution of random walks.

TABLE VII. PRIMARY FEATURES OF MODELS OF GRAPH REINFORCEMENT LEARNING

| Models | Task | Actions | Rewards |
|---|---|---|---|
| GCPN [132] | Graph-generation | Prediction of links | domain knowledge + GAN |
| morgan [124] | Graph- Generation | Generation whole graph | domain knowledge + GAN |
| GTPN [125] | Chemical-reaction | Prediction of bond types & node-pairs | Resultant Prediction |
| GAM [126] | Graph-classification | Prediction of graph labels & next-node selection | Resultant Classification |
| DeepPath [127] | KG-reasoning | Prediction of next- node of reasoning-path | Diversity + resultant reasoning |
| MINERVA (121) | KG-reasoning | Prediction of next- node of reasoning-path | Resultant Reasoning |

TABLE VIII. MAIN PROPERTIES OF GRAPH ADVERSARIAL MODELS

| Class | Model | Adversarial method | Node Attributes |
|---|---|---|---|
| **Adversarial Training** | ARGA/ARVGA [133] | GAEs regularization | Yes |
| | NetKA [134] | GAEs regularization | No |
| | GCPN [122] | Graph RI rewards | Yes |
| | MORGAN [124] | Graph RI rewards | Yes |
| | GraphGAN [135] | Negative sample generation | No |
| | | Network-embedding regularization | No |
| | GraphSGAN [136] | Enhancing semi-supervised learning on graphs | Yes |
| | Netgear [115] | Graph-generation via random-walks | No |
| **Adversarial Attack** | Attack [110] | Targeted attacks on graph structures & node-attributes | Yes |
| | Dai et al. [137] | Targeted-attacks of graph-structures | No |
| | Zugner &Gunnemann [115] | Non-targeted attacks on graph-structures | No |

## B. Adversarial Attacks

The goal of adversarial attacks that belong to the category of adversarial approaches is to purposefully "fool" targeted strategies via appending slight perturbation to data—analysing adversarial assaults aid in understanding current models better & lead to more durable designs. Below is a brief analysis of graph-based adversarial assaults.

Attacking node classification models like GCNs was first suggested by Nettack [110] by altering graph architecture &node properties. The target model is denoted as $\mathcal{F}(\mathbf{A}, \mathbf{F}^V)$, the loss function as $\mathcal{L}_\mathcal{F}(\mathbf{A}, \mathbf{F}^V)$, the target node as $v_0$, &the true class as $c_{true}$. The model adopted the following objective function:

$$\underset{(\mathbf{A}', \mathbf{F}^{V'}) \in \mathcal{P}}{\operatorname{argmax}} \max_{c \neq c_{true}} \log Z^*_{v_0,c} - \log Z^*_{v_0,c_{true}} \qquad (68)$$

$\text{s.t. } \mathbf{Z}^* = \mathcal{F}_{\theta^*}(\mathbf{A}', \mathbf{F}^{V'}), \theta^* = \operatorname{argmin}_\theta \mathcal{L}_\mathcal{F}(\mathbf{A}', \mathbf{F}^{V'})$. Where $\mathbf{Z}$ stands for categorisation probabilities estimated through $\mathbf{F}()$, $\mathbf{A}'$ & $\mathbf{F}^{V'}$ denotes modified adjacency matrices & node attributes, respectively, *and P* denotes space allotted via attack restrictions. Essentially, optimisation seeks to identify the most appropriate modifications to graph topologies &node properties that will lead to the misclassification of $v_0$. $\theta^*$ shows that the attack is causal, i.e., before the targeted model is trained. An attack must only cause minor modifications to be "unnoticeable," which is the main restriction apart from preserving data properties like distribution of node degrees & co-occurrence of features. Many relaxations & attacking scenarios—*directly targeting $v_0$* & only attacking other nodes *(influence attack)* make optimisation manageable. Reference [138] proposed graph-adversarial attacks with a target function equivalent to the above Eq., especially in scenarios when only graph structures were altered. The most effective technique, RL-S2V, learned node & graph representations using structure2vec [50] & solved optimisation using reinforcement learning suitable for both graph & node level classification activities. as above- mentioned methods induce misclassification of a target node $v_0$. Reference [110] studied *non-targeted assaults* to lower the model's overall performance. They employed meta-gradients for optimisation & considered graph structure as hyper-parameters to be enhanced.

## VIII. GENERAL APPLICATIONS

GNNs are employed on graphs (non-Euclidean data modelled through graphs), whereas classical neural-networks work on arrays. Graphs have been extremely popular in recent years as a result of their capacity to depict real-world problems in an interconnected manner. The data in the applications are structured. However, data is utilised in unstructured forms, such as images modelled as graphs for analysis in social networks, chemical structures, and web-link data.GNNs offer a wide range of utilities in a variety of application areas. Although activity tasks such as node or graph classifications, graph generation, specifically each GNN class, handles network embedding& spatial-temporal graph forecasting.

There are three Taxonomical Classifications of general applications of GNN, which are based on:(1) **the Nature of Data**, i.e. Structured/Unstructured Data, (2) **Learning mode** (supervised/unsupervised), which is further depicted by primary Task to be learned (i.e. Edge Prediction, Node or Graph classification) after modelling actual problem into graph form & (3) **Field of Study** where GNN is to be applied. We present a Taxonomical Classification of applications of GNN in Real-life scenarios based simultaneously on *the nature of data* & *Field of Study* where to be applied, as is depicted in Fig.13.

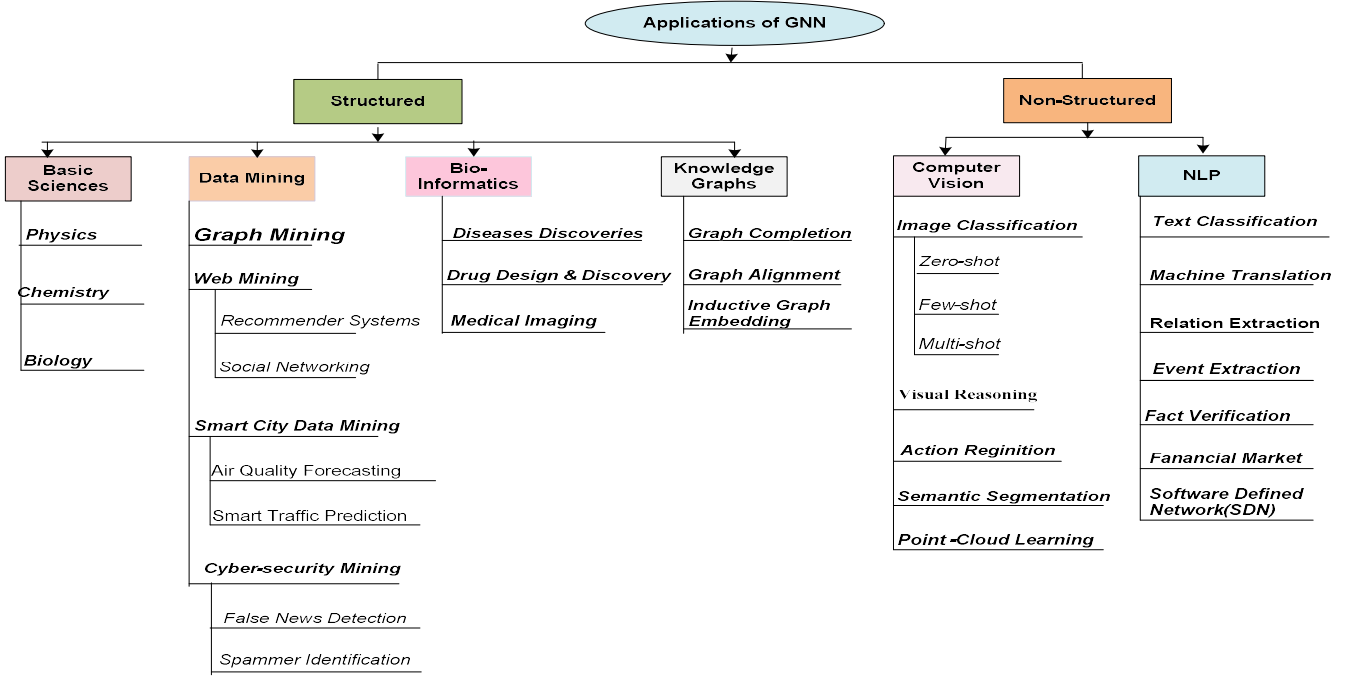

Fig. 13. Taxonomy of Applications of Graph Neural Network *(based on Nature of Data & Field of Study)*

TABLE IX. APPLICATIONS OF GNNs IN DATA MINING

| Applications | | Description |
|---|---|---|
| **Graph Mining** Extracting valuable structures for later tasks. | **Graph Matching** Detecting similarities between graphs is known as graph matching. Used in pattern recognition & computer vision to encode structural information | GNNs capture the structure of graphs. [139] suggest a model comprises of two parallel MPNNs that share a similar structure & weights to implement a graph pair with minimum editing distance into local latent space.[140] develop comparable strategies while conducting trials control flow graph similarity search. |
| | **Graph Clustering** It is the process of clustering a graph's vertices into clusters according to the topology of the graph & attributes of nodes. | Classic clustering techniques [141] for the representation of nodes. Graph pooling [142]- clustering goes beyond learning node embeddings. [143] improve the spectral modularity metric in graph clustering. |
| **Web Mining** Finding patterns in Web-based application data. | **Recommender Systems** A recommender system is a form of information filtering system that promotes objects based on what customers can find most valuable. To combat the challenge of information overload, this technology is widely used in online businesses such as social networking, video/music streaming solutions & e-commerce. | The collaborative filtering approach uses users' previous behaviour data to forecast their preferences [144]. [145] employ collaborative filtering to encode information into vector representations of entities that reconstruct previous interactions. |
| | | The key to Graph-based recommender systems is to assess whether an article is relevant to a customer. So, it can be designed as a problem in predicting relationships. [146] GAE uses a convolution GNN as an encoder to predict missing links between users & objects. |
| | | RNNs with graph convolution are integrated by [147] to understand how known scores are generated. |
| | | PinSage [142] constructs computational graphs for bipartite graphs through a weighted sampling technique to reduce redundant processing. |
| | **Social Networking** Users in social networks are represented via nodes & interactions/relationships through the edges; hence social networks can be readily described through graphs, e.g. Facebook users may indicate their interests in various movies, sports, & literature described as graphs. | GNNs aid in social impact prediction [148], political perspective recognition [140], and social representation learning on social networks [102] to estimate regional social impact for user participants in social networks[148], including capturing preference similarities among users in different behaviours to learn corresponding user's representations [102] |
| | | User embeddings are learned by GraphRec [149] from both the item & user sides to improve suggestion performance. |
| | | [150] extend suggestion performance beyond static social impacts by using dual attentions to model homophily & influence effects & the bipartite graph to reduce repeated computation. |
| **Smart City Data Mining** | **Air Quality Forecasting** | The air quality forecasting issue may be modelled through spatial-temporal GNNs as air quality at neighbouring areas is connected & the air quality in one site is temporally changing. |
| | **Smart Traffic Prediction** Accurately estimating traffic volume, speed or density of highways in traffic networks is critical. | STGNNs solve the traffic prediction problem in [32] [20] & [8] by modelling the traffic as the spatio-temporal graph, with nodes representing road sensors, edges representing the distance between nodes & dynamic input characteristics representing the average traffic speed. |
| | **Taxi Demand Forecasting** Based on weather data, past taxi demand, geo-location data & event considerations. | LSTM, CNN & network embedding are trained via LINE [97] to construct a unified depiction to estimate the count of cabs requested for the location within an interval of time [9]. Attention mechanisms by [151] & [152] better model spatio-temporal correlation. |
| **Cyber-Security Mining** | **Spammer Identification** | GNNs tackle cyber-security data issues such as spammer identification & false news detection [153], [154] as cyber security data can be represented as graphs. |
| | **False News Detection** | |

## A. Structural Scenario Data Applications

Structured Data in GNN stands for the data which can naturally/explicitly cast in graphical/relational representations, e.g. traffic prediction, recommender systems, graph representation, modelling physical systems, forecast molecular chemical characteristics & protein biological interface features. Following is a brief overview of structured data-based real-life applications of GNN.

### 1) Data Mining

The data-mining goal is discovering patterns and information out of enormous volumes of data. Multitudes of real-world data can be presented via graphs, e.g. Social graphs are web-based representations of relationships between social media users. GNNs can be used for various data mining activities, as depicted in **Table IX**.

### 2) Knowledge Graphs (KG)

Knowledge graphs (KG) are representations of real entities in addition to the relationship that occurs across them and are widely used for different applications, including information extraction and question-answering. It can further be utilised under the sub-domains of KGs as described in **Table X**.

### 3) Basic Sciences

#### a) Physics

The components of a system, along with pair-wise interaction, represent any **Physical System**. Simulation demands the model learning the system's law & making predictions about the system's next state. Systems as depicted in Table XI. can be simplified as graphs by describing the items as nodes & pair-wise interactions as edges

#### b) Chemistry

A molecular fingerprint is a way of encoding a molecule's structure. One-hot vector is the simplest fingerprint, with each bit reflecting the existence or absence of a given substructure. Researchers utilise GNNs in chemistry to model molecules where edges denote bonds & nodes depict atoms, as specified in Table XI.

#### c) Biology

Models of Graph neural networks used in a diverse range of biochemistry & healthcare services such as drug discovery & development, side effect prediction of polypharmacy, multi-view drug resemblance integration, sickness prediction relationship & drug recommendation. For more details, please refer the section IX .

TABLE X. APPLICATIONS OF GNNs IN KNOWLEDGE GRAPH

| Sub-Domains of KG | Description |
|---|---|
| **Knowledge Graph Completion** Process of deriving missing information from data already present in a knowledge graph & suggests several forms of error functions to distinguish real triples from erroneous triples. | R-GCN provides apparent connection modification in message forwarding phases to deal with varied relations for knowledge graph embedding [28]. |
| | Structure-aware Convolution Network integrates GCN and CNN as encoder & decoder, respectively towards the excellent presentation of knowledgebase [155]. |
| **Inductive Knowledge Graph Embedding** Completing knowledge bases for **OOKB** (Out-Of-Knowledge-Base) Instances. | An OOKB instance does not exist in the training dataset but is associated with observed instances. Observed entities are used to aggregate the embeddings of OOKB entities by using GNNs in both conventional KBC & OOKB settings [156]. |
| **KG Alignment** KG alignment matches entities in different KGs, which is crucial for knowledge fusion and handling heterogeneity challenges. Combines elements from diverse languages into a single embedding & conforms them depending on similarities in embedding space. | GCN tackle the cross-lingual graph-based knowledge alignment challenge [133] |
| | Graph attention networks represent entities for aligning large heterogeneous knowledge graphs[141]. |
| | The Entity alignment Challenge is converted to a problem of graph matching & using graph matching networks for modelling objects as their adjacent sub-graph [157]. |

TABLE XI. APPLICATIONS OF GNNs IN PHYSICS

| Application | Description |
|---|---|
| **Particle Systems** | Interaction of particles with one another, like **collisions** [41], and electromagnetic force [116], can be represented through graphs. |
| **Robotic Systems** | Graph Network-based approach [158] for encoding the graph formed for a robot & combine GNs with Reinforcement learning to acquire the policy of reliably managing the system. |
| **Dynamic Modelling** | NRI [116]takes object trajectory as input & derives an expressive interaction graph while simultaneously learning a dynamic model. Graph Interactions are learned via previous trajectories and also interpreted for trajectory estimations. |

TABLE XII. APPLICATIONS OF GNNs IN CHEMISTRY

| Application | Description |
|---|---|
| **Molecular Fingerprints** | Reference [3] present neural-graph fingerprints(Neural FPs), employing GCNs to build feature-vectors of substructures & then aggregate them to provide overall representations. |
| **Chemical Rxn-by-product** | Graph Transformation Policy Network [125] encodes input molecules & generates an intervening graph that includes a vertex pair estimation graph & a policy network. |
| **Atom interactions** | [38] explicitly simulate atoms & atom pairs separately. |

TABLE XIII.   APPLICATIONS OF GNNs IN COMPUTER VISION

| *CV Application* | | *Description* |
|---|---|---|
| **Image classification** ImageNet datasets [160] | **Zero-shot:** Only *N=0* subsamples in the same classes are offered in training phase to anticipate for the test data samples. | Knowledge graphs(KGs) are employed to supplement zero-shot recognition categorisation [133] and train visual classifiers (KG encoded with 6-layer GCN) to learn from visual input, word embeddings of class names & links to other classes. To avoid the over-smoothing effect,[161] used a single GCN layer with a wider neighbourhood that contains both one & multi-hop vertices in the graph. |
| | | [5],[162] choose associated objects to generate a sub-graph depending on the outcome of the recognition of the object & employ GGNN to them for determination since most information graphs are too big for reasoning. |
| | | [163] use knowledge networks between categories as well. It defines three forms of inter-category relationships and directly disseminates the certainty of relationship tags in the graph—the resemblance of images within the dataset aid in few-shot learning [164]. |
| | **Few-shot** | Few such training instances from equivalent classes are offered in training to predict for test data samples. [164] develop a fully connected weighted image approach that relies on likeness & executes message passing via graph. |
| | **Multi-shot** | Sufficient samples from the same classes are used in training to predict test data samples. |
| **Visual Reasoning:** It refers to the ability to comprehend behaviours & logic connected to any visual images.CV systems do reasoning based on spatial & semantic data. So, generating graphs for reasoning tasks is natural. | | **Visual Question Answering (VQA)** A CV representation must respond to inquiries regarding an image depending on the question's wording descriptions & answer is found in spatial relationships among entities in the image. [165] design a query syntactic graph & an image scene graph. The embeddings are then trained using GGNN to estimate the final result. [166] construct relational graphs based on queries. [133] & [167] used KGs to do deeper relation research & more interpretable reasoning. |
| | | GNNs are utilised to calculate RoI features for **object detection** [132], [168], as messages-passing tools between humans and objects in **interaction detection** [169] & execute reasoning on the graphs that link classes and regions for the **region classification** [170]. |
| **Semantic Segmentation:** It refers to a mechanism for labelling each pixel in a picture uniquely. Segments in images are not grid-like & require non-local details, causing classic CNN to fail & graph-structured data to come to the rescue. | | **Scene Generation** earning underlying a visual sight is made more accessible by recognising Semantic links between items. Models for scene graph design strive to interpret images together into semantic graph networks of entities & associated semantic interconnections [137], [138], [139]. However, one application [140] reverses the procedure using scene graphs to generate realistic images. Since natural language could well be interpreted into semantic networks, with each word/term denoting an item, it is a viable approach to build visuals from the descriptive text. |
| | | **Point-Cloud Classification:** Point-Cloud specifies a collection of 3-dimensional point-coordinates formed via LiDAR scanning.[141], [142], [143] employ ConvGNNs to assess the topological information of the Point Clouds converted to k-nearest neighbour graphs. |
| | | **Action Recognition** Identifying human behaviours in video footage, specific solutions recognise locations of human joints joined via skeletons, producing a graph.[73], [75] employ STGNNs to learn activity patterns from time-based sequences of human joint coordinates. |

B. NON-STRUCTURED SCENARIO DATA APPLICATIONS

Un-Structured Data Scenarios represent the data where the graphical/relational structures are hidden or missing, e.g. **image, text**, multi-agent systems & programming code. On non-structural contexts, there are two approaches for using graph neural networks: (1) incorporate topological information across other realms to boost efficiency, such as utilising knowledge-graphs for relieving zero-shot concerns in operations of an image; & (2) presume structural- relationship in a specific situation & employing GNN approach to solve concerns outlined on graphs, e.g. the approach employed by [159]. Detailed

overview of structured data based real-life applications of GNN is given as under:

**1) Computer Vision (CV)**

It is the subfield of artificial intelligence, which facilitates machines to extract relevant data out of movies, photos or other inputs, as well as perform tasks or make suggestions based on this data. Table XIII describes the various applications of GNNs in different CV tasks.

**1) Natural language processing**

NLP (Natural-Language Processing) specifies a discipline of "artificial intelligence" (AI) which allows computer-systems to interpret written & spoken language in a way analogous to human-beings. GNNs might be used for a variety of text-based applications at both word-level tasks & sentence-level tasks as depicted in **Table XIV**.

TABLE XIV. APPLICATIONS OF GNNs IN NATURAL LANGUAGE PROCESSING

| Application | Description or Mechanism Involved |
|---|---|
| **Application** **Text Classification** (e.g. Bag of Words) | Text as a representative graph includes words that derive interpretations correctly across long-distance & non-consecutive words by a **Deep graph-CNN model** [171] and then convolves the word graph using graph convolution procedures. |
| | [172] present the Sentence-LSTM for encoding text data by assuming the complete sentence as one specific state, with sub-states for each word & a sentence-based state. |
| **Sequence Labelling** Assigning a label class to each word in a sequence of observed words. | **POS-Tagging** identifies the words present in a sentence according to its parts of speech. **Named Entity Recognition (NER)** predicts if every word contained in a sentence is an element of an effective entity. Sentence LSTM is used by [172] to label the sequence where they experimented with POS-tagging & NER tasks & have had good results. |
| | **Transformers** ([30] replaces recurrent or convolution layers with attention methods assuming words are coupled in a completely connected graph structure. |
| **Neural Machine Translation** Sequence-to-sequence task that automatically led to text translation across different languages via neural networks | [173] & [174] use **Syntactic GCN** for performing syntax-aware Neural Machine Translational (NMT) operations & the latter extends to include information on the predicate-argument pattern for source phrases [175] use the GGN for embedding's edge labels through various transformations. |
| | Traditional involves methods like **CNNs or RNNs**, while sentence dependence structure is a more advanced method for learning & predicting relations. [172] present a relation extraction extension of graph convolution networks that uses a pruning technique on input trees. |
| **Relation Extraction** Extraction of semantic relationships between entities in texts extends the knowledge base. | [171] use **Graph LSTMs** for N-ary relationship retrieval discovers relationships among n entities across several sentences. [176] also, employ the graph-state-based LSTM model to boost parallelisation & speed-up processing. |
| | [177] & [178] utilised **Syntactic GCN** dependency trees for event detection |
| **Event Extraction** Identifies various events in texts by identifying the event triggers & predicting arguments | Syntactic shortcut edges promote the flow of information to **attention-based GCNs** to represent graph information, [179] provides a **Jointly Multiple Events Extraction (JMEE)** paradigm to retrieve various event causes & descriptions simultaneously. |
| | Evidence aggregation & reasoning relying on the fully-connected knowledge graph are proposed using **GNN-based approaches** such as GEAR [146] & KGAT [179]. Inner-sentence graph is formed using the knowledge from semantic role labelling & achieved encouraging results [180]. |
| **Fact Verification & others** Verification of natural text facts against a database of facts uses models to retrieve justification to validate specified facts. | **QA & Reading** GNNs can also be used for various other text-based activities. [181], [182], [183], [136] apply GNNs in question answering & reading comprehension. |
| | **Interaction Networks:** Another central direction is relational reasoning tasks based on text, interaction networks [52], recurrent relational networks [184] & relational networks [56] have been developed. |
| | GNNs are also employed in the financial sector to predict the interaction between multiple stocks to manage plans. [136], [170], [179] [185]. |
| | GNNs is supposed to enhance performance for Network routing tasks, as in [186]. |

| **Finance Market Prediction** | It disassociates from syntactic representations --sentences related in meaning should be assigned the exact Abstract Meaning Representation (AMR), even if not similarly worded. [181] [175] employ GNNs to encode the graph representation in AMR graph to Text production challenges. |
|---|---|
| **AMR Graph-based model which preserve semantic relationships among words in text** | Evidence aggregation & reasoning relying on the fully-connected knowledge graph are proposed using **GNN-based approaches** such as GEAR [146] & KGAT [179]. Inner-sentence graph is formed using the knowledge from semantic role labelling & achieved encouraging results [180]. |

TABLE XV. APPLICATIONS OF GNNs IN GRAPH GENERATION

| Models | Description | References |
|---|---|---|
| **Netgear** | Generates adjacency matrix using GraphRNN | [115] |
|  | Sequential generation of edges & nodes using GNN | [179] |
| **GraphAF** | Blends autoregressive model with the flow-based generation with validation check after each phase of molecule production | [187] |
| **MolGAN** | It employs a reward network for RL-based chemical property optimisation & solves the node variant issue with a permutation-invariant discriminator. | [182] |
| **GCPN** | It also uses reinforcement learning to incorporate domain-specific rules. | [122] |
| **GNF** | The generative model uses invertible mapping for encoding & inverse-matrix for decoding. | [130] |
| **GNN with AE** | Encode the graph structure & characteristics into latent variables using isotropic Gaussian & decoding is done using an iterative refining technique. | [188] |

TABLE XVI. APPLICATIONS OF GNNs IN COMBINATORIAL OPTIMIZATION

| Optimisation Application | Description |
|---|---|
| *TSP & MST* | GNNs are an effective way to enhance the conventional approaches, as mentioned in [51] & [189]. The node embeddings are obtained from structure2vec [50] & then transferred to a Q-learning-based reinforcement system for making decisions which construct an encoder-decoder system based on attention demonstrating the representation capability of GNN. |
| *Bi-partite graph* | [190] use GCN to encode the state of a combinatorial puzzle, which they describe as a bipartite graph. |
| *Quadratic Assignment Problem* | It involves assessing of similarity of two graphs for specific combinatorial optimisation issues. The model based on GNN learns node embeddings for every graph separately & uses an attention method to match them [87]. |
| *DAG Learning Problem* | It is both the NP-hard & combinatorial optimisation problem [151] use a generative graph neural network |
| *SAT Satisfiability* | NeuroSAT [191] develops a neural network based on message passing for the classification of satisfiability of SAT problems where a learnt model can be applied to new SAT distributions & other problems convertible to SAT. |
| *Generalised Optimisation* | [192] provide a conceptual evaluation of GNN models on the challenges mentioned above & provides links between GNNs & collection of traditional graph algorithms for handling these problems. It shows that the most powerful GNN can achieve the best approximation ratios (by adding colours to the node feature) to the best solutions. |

## C. OTHER APPLICATIONS OF GNN
### 1) Graph Generation
Tasks like social interaction modelling, exploring new chemical structures and generating knowledgebase graphs are just a few of the uses of generative models in real-world graphs. There has been a recent increase in neural graph generative models due to the exceptional power of deep learning approaches to understand graphs' latent distribution, as described in **Table XV**.
### 2) Combinatorial Optimization
Problems involving optimisation via combinatorial means on graphs are a collection of NP-hard challenges. Heuristic solutions exist for situations like the minimal spanning trees (MST) & travelling salesman problem (TSP). Recently, employing deep neural networks to solve such challenges has become popular & several of these solutions use neural graph networks, as mentioned in **Table XVI**, due to their graph structure.

# X. BIOINFORMATIC APPLICATIONS

In this section, various applications of Graph Neural Networks Bio-informatics are discussed. Based on multiple biological tasks, GNN's present applications in bioinformatics can be divided into three key aspects: Drug Design, Disease-Related Variable Discoveries & Medical Imaging.

## A. Disease-Related Factor Discoveries

Predicting factors associated with a variety of Diseases is a significant task in bioinformatics. Various existing methods currently being used for predicting disease association are matrix degradation [169], network propagation ([193][146] & machine learning [194], [195], [196]. However, the machine learning-based methods-- shallow models (e.g. matrix factorisation) cannot depict sufficient structural knowledge in disease-related graph networks, which influences feature modelling effectiveness. GNNs have subsequently been utilised in biological networks to capture non-linear interactions between diseases & other elements. Convolution procedures are increasingly used in heterogeneous networks to extract local sub-graph features. Disease-RNA relationships, gene-disease associations & various critical data-related biological networks were explored, which led to the invention of deep learning algorithms for disease prediction, as illustrated below:

### 1) RNA --Disease Association

According to Various studies [151], [195][151], [195] Circular RNAs, Piwi-interacting RNAs, MicroRNAs (miRNAs) & Long Non-Coding RNAs (lncRNAs) are involved in the onset & progression of the disease. As a result, determining the association between the different RNA --diseases is crucial in understanding the origin of complex diseases. These methods are computationally intensive but still make up for the cost & time-consuming faults of formal verifications of bio-experiment. GNN-based approaches like [197] offered the DimiG, a multi-label semi-supervised graph convolution model that does not depend on known disease-miRNA relationships. Network information sharing is used in DimiG to estimate the connection between miRNAs & disease by combining several networks connected with genes expressing proteins. GCN is used by [159] to learn feature embeddings of diseases & miRNAs via disease semantic similarity network & miRNA functional similarity network, respectively. This GCN can also generate an association matrix using the neural induction matrix, integrating known disease & miRNAs connection information into the model's training. The model can identify all miRNAs tied to breast tumours without any known associated miRNAs. Prediction of associations between RNA & disease can combine additional valuable data based on methods of similarity. PPI networks, miRNA-gene & disease-gene are integrated by Li C et al. (2019). In addition, the top ten unknown relationships between disease & miRNAs were analysed based on information extracted from GCN. [129] utilising the FastGCN method & the classifier Forest to accurately predict the association of potential circular RNA diseases by penalising attributes.[198] GCN was used in tandem with attention mechanisms to derive domain properties & also experiments were conducted on two RNA disease networks to learn the latent representation of node functioning better.

Some studies employ the auto-encoder approach on the network to recover the node properties. [134] applied GCN-based encoder to retrieve lncRNA & disease characterisation from a bipartite graph connected with lncRNA disease & its corresponding interaction score are two possible variables determined by operating the vectors on the inner product. In the study by [199], **VGAE** minimises the noise generated by arbitrary negative instances. Integrating multiview information can assist in understanding complicated biological networks for detecting Diseases-associated RNA with little labelled data. Hence, additional study is required on integrating different data types with graph deep learning models to comprehend deeper modes of interaction between multiple related data.

### 2) Relationship between diseases & genes

Technology for RNA sequencing of single cells provides insights into single-cell gene expression. Accurately predicting the association between a gene & a disease helps researchers clarify the function of the gene responsible for the disease & offers proof of disease prevention. Emphasising candidate genes for diverse diseases can hasten the discovery of early-stage medications & overcome specific DGP concerns. GNN presumes cell-cell interactions [135], [200], [201], models cell differentiation (Bica et al., 2020) & sickness can be diagnosed [202]. Reference [203] developed a novel method for sickness gene sequencing that varies from the general interaction network approach. It further expressed pairwise negotiated and semantic relationships into a heterogeneous graph. & utilised frequencies qualifier from the **Orphanet** to estimate the edge weights [204]. Since standard VGAE is not able to learn the connections between various node kinds. [205] introduced a **restricted VGAE version** for estimating specific node connections on the gene-disease interaction network via updating the algorithm's optimising goals.[206] established a novel dropout method & cluster error function that relies on graph embedded approach & GCN to boost the generalisation.

Multiple gene defects cause cancer as a set of complicated disorders & there is synthetic lethality (a circumstance in which mutations in two genes jointly result in cell death) between genes. Thus, interaction networks of genes play a significant role in the diagnosis of cancer [121]. In order to examine the biological mechanisms of the tumour, [207] first employed GCN to categorise tumour genes. Layer-wise relevance propagation is utilised to detect the gene data input & network structure of the learnt model adjacent to a gene. As Synthesised toxicity across genes is incredibly low, [208] approximated this by presenting a novel GCN model based on coarse-grained node dropouts & fine-grained edge dropouts to minimise the overfitting of sparse networks.Reference[209] integrated patient gene expression data with the PPI network & utilised GCN to categorise nodes inside the patient's subnet to detect breast cancer metastasis. Relevant research exists utilising local GCN to aggregate information from gene expression matrices & the PPI network in several patients to categorise tumour subtypes [210]. The GNN-generated correlations for every dataset not only enhance the comprehensibility of the model but also assist in predicting tasks associated with the patient-specific disease networks

### 3) Discovery of Disease Proteins

GNN has also been introduced in related studies, such as **disease discovery** [211]. **Detecting pathogenic proteins** can be treated as semi-supervised classification by employing **a protein-protein interaction** network. The realisation of neighbourhood positioning of the visualised disease pathway indicates that most diseases lack prominent neighbourhood positioning. Graph structures can model secondary RNA molecules for RNA classification [126]. They are also used to predict RNA-binding proteins [212], where bases are represented as nodes & hydrogen bonds & phosphodiester bonds are two different kinds of edges. Some research has employed lncRNAs, miRNAs & other components to develop heterologous networks for predicting interactions between these RNAs & genes that target lncRNAs. Based on various omics (analysing the interactions & functions of biological information entities in various clusters of life) data & cosine similarity techniques to classify biomedical data into multiple groups weighted patient similarity network is constructed [129]. GCN conducts extraction of features on these networks to reveal multi-omics correlations within labelling space & integrate multi-omics pharmaceutical research & discovery successfully.

## B. Drug Development & Discovery

The **development of drug**s primarily includes drug discovery target identification, lead molecule discovery & optimisation, drug prospect assessment, and preclinical & clinical investigations [213]. However, the absence of new therapeutic targets, poor clinical translation of models of animals, heterogeneity of ailments & the inherent biological system complexities has made drug development a lengthy, complex procedure. The objective of current drug design is to hasten & improve intermediate phases utilising machine learning approaches to minimise developmental expenses to forecast preliminary molecular features, which may decrease the labour of subsequent trials. Deep learning is the ideal machine learning approach in bio-medicine. However, it has two primary limitations: its techniques cannot directly learn information about the structure from input source data, so they depend on highly labelled quality data. Also, Classical CNN needs help directly processing directly unstructured data, e.g. molecular graphs, since the intrinsic structural features of compounds are lost. As a result, GNNs have emerged as the latest methods for dealing with drug-related activities (like molecular representation learning, molecule graph generation, and drug-target binding affinity prediction), as discussed below.

### 1) Protein Structure & Functional Prediction

Protein function research is crucial in the life sciences, as protein malfunction is linked to most disorders.[214] discovered that a denatured ribonuclease with merely the basic structure may refold & recover biological activity, demonstrating that the amino acid sequence encoding of protein's primary structure provides essential details about secondary & tertiary structure. On the molecular scale, the most accurate Protein structure forecast can thoroughly explain the bio-mechanism of protein functioning, & its usage in drug development & other sectors is vital to biochemical study. High computation cost & comprehensibility are concerned with traditional techniques of molecular analysis of structures like 3D-CNN & 2D-CNN.

Recent research has revealed GNN's compelling ability to learn effective structures of proteins from simple representations of graphs and [215] introduced an excellent protein structure learning strategy for large datasets, which can be applied to realistic 3D representations of molecules, providing a high level of transferability in the field of applications. Reference [218], addressing the challenge of inverse protein folding, presented the conceptual design of protein relying on related graph-attention strategies, which can lead to a conditional generative model of a key target sequence of the protein directly and significantly improve functionality. There are two methods for predicting protein function: protein structure-based methods & PPI network-based methods. Reference [216] depicted Protein structure as the graph for predicting the function of the protein. Reference [217] employed a multi-relational graph approach relying on PPI network modelling via semi-supervised learning. Many invalid segments in a protein sequence may impact the evaluation of its functionality. Integrating the functionality of protein-relational networks using GNN is one way to solve protein sequence & functional differences.

### 2) Prediction of protein-protein interactions

Proteins are chains of organic compounds, i.e. amino acids, with biological activities [218]. It has carboxyl (-COOH) & amine (-NH2) functional groups, & a distinct side chain (R group) for each amino acid. Proteins must interact with other proteins in order to carry out their activities. Predicting the interface at which these interactions take place is a daunting challenge with substantial significance for drug research & design. The protein–protein interface comprises interacting amino acid residues and neighbouring amino acid residues in interacting proteins. As per [218], two amino acid residues from distinguished proteins are recognised to constitute an element of an interface when any arbitrary non-hydrogen atom in a residue of amino acid is within 6°A for any arbitrary non-hydrogen atom in another amino acid residue. Therefore, protein interface prediction could be addressed as a **binary classification task**, with input as pair of amino-acid residues representing different proteins. Protein residues amino-acid are expressed as vertices & relations are expressed as edges in the graph. The node representations are subsequently learned using graph neural network models, which are then used for classification.[219] suggested a spatial convolution operator for forecasting the interface across protein pairings for graphs of variable size & structure. The PPI network's high false positive/negative rates make it challenging to recognise protein structures. Consequently, [9] suggested a noise reduction approach based on an Autoencoder for fluctuation graphs. They embedded a PPI network in feature space via a multilayer GCN & eliminated specific sub-threshold reliability interactions to yield a reliable association network. Extensive experiments on numerous datasets have revealed that the identification accuracy of protein structures is enhanced by 5 to 200%. **Unsupervised GNNs** [220] were used to forecast changes in the protein's binding characteristics following mutation & to find anomalous interactions between atoms without annotation. References [221], [222] established a scoring method for evaluating **Protein Docking Models** & treated peptides, respectively, by enhancing existing sorting algorithms. The graph convolution operation encodes protein structures & features into graph-embedded representations that combine information all along edges connecting network nodes to evaluate relevance & address the spatial limits of traditional convolution techniques.

### 3) Drug-Target Interaction Estimation

Medicine targets are associated with the pathological state of the biomolecule or diseases. So, identifying drugs (ligands) & their targets(proteins) is a critical task in novel drug development. Novel drug development is typically time-consuming & costly. Drug-target interactions (DTI) must be detected early in drug development to reduce the candidate medication search space. Also, it is utilised for drug repurposing, which tries to find novel applications for outdated or unused medications. Identifying drug-targeted interactions is a matter of predicting relationships involving proteins & ligands, as numerous relevant research has been undertaken. However, challenges remain, such as (1) Conventional machine learning approaches represent prediction outputs via binary classification, although the actual association is more expansive than a binary level. The prediction accuracy cannot be obtained in case the target proteins are nonexistent in the test dataset but present in the actual applications. (2) Difficulty handling the molecular space in which drug compounds can be generated. (3) The test findings for the model appear decent. However, they must be convincing (4). Data on the relationship between truly negative ligands & proteins for learning needs to be improved. GCN was introduced firstly in **drug target identification** which learnt drug molecular structure information & coupled protein information as input to address these challenges [223]. This study solves the drug-target dilemma by predicting the true interaction strength between medications & targets. Several other studies [122],[177] are similar to the already stated approach's fundamental idea but vary in data processing.

Reference [224] employed an unsupervised graph auto-encoder to generate the depiction of protein pockets independent of the target-ligand complexity applying features derived by GCN from pocket graphs & 2D ligand graphs, respectively. [225] provided a model for predicting drug-target interactions in which ligands specifically target proteins without using actual negative interaction information. [196] proposed a related prediction method for constructing molecular graphs & protein contact graphs. These figures are derived from structural information on medicinal compounds & sequence information on proteins synthesised via the 3-layer GCN to provide accurate predictions [226], as opposed to the methods cited above to model drugs as graphs suggested a graph-based estimation representation that included drug-protein association network in existing approaches. The data feature learnt by GCN for each drug-protein combination is sent through a deep neural network to determine the output label. [180] employed gene transcriptional patterns to estimate drug-targeted binding indirectly to compensate for earlier studies' lack of knowledge of Cellular Context-Dependent Effects. The reliability of the **graph databases** in GNN for the drug was closely linked to the outcome of all graph-based association predictions in these investigations. However, the nature of graph databases obtained in real life is complex & involves overlapping information. The implications of noisy data upon prediction accuracy of the model are reduced by [227] allocating small weights to uneven edges.

### 4) Molecular Properties Estimation

Deep learning techniques were used to make predictions regarding the properties of chemicals using molecular data. As molecules can be of any size or structure, deep learning approaches like convolutional neural & feed-forward could not be implemented on molecular data directly. So, firstly features are extracted as molecular fingerprints in vector form depicting the molecule's structural information & then properties are predicted using the same through deep learning algorithms. Traditionally, the molecular fingerprint has been recovered using non- differentiable off-the-shelf fingerprint software. As a result, these extracted representations may not be ideal for prediction.

Initially, graph convolution in molecular properties learning [3] relied on extended connectivity circular fingerprints (ECFP). It built differentiable patterns to substitute discrete functions in circular patterns & the hash functions were exchanged with one-layer neural networks. Experiment conclusions indicated that this model performed similarly to ECFP using random & large weights

& they performed better than ECFP when the weights were adjusted during training.

Reference [38] proposed a chemical graph convolution approach using deep neural networks rather than molecular fingerprints that otherwise lack noise & space parameters. A molecular graph represents the molecular structure & the in-between graph distance determines the level of the molecule. This approach is sometimes preferable to the chemical fingerprint procedure. In addition, with the advent of **multitasking deep neural networks** (MTDNN), the neural network has become more potent in drug discovery. [228] coupled GCN & MTDNN to enhance prediction performance further, resulting in a complete knowledge-driven deep learning approach that does not depend on context-specific attribute descriptions or biomarkers for estimating drug attributes. Generally, electrostatic computations help predict the chemical reactivity of compounds & their capability to make specific interactions.

Reference [229] developed a technique to build an electrostatic potential surface near the quantum mechanical quality of both ligand & molecule inside the timespan of active drug design. It offers an effective solution for drug design. In the drug discovery mechanism, the findings of false-negative or false-positive experiments induced with unstable molecules in storage make it hard to validate estimates of chemical stability. [183] proposed an Attention-GCN model to predict bond stability. This model proactively learnt spatial features from the molecular network rather than established structural data, lowering the probability of false positives. Graph convolution procedures would extract local aspects of chemical sub-structure interactions & provide universal descriptors of mixed structure data [230].

**Novel Molecule Design**
Novel /De novo Molecular Design (DNDD) is regarded as the development of new chemical compounds utilising computational growth algorithms that meet a set of criteria. The phrase "de novo" denotes "from the beginning", indicating unique molecular units without a starting layout. The main objective of drug design is the discovery of molecules with ideal chemistry. **Computational drug design** helps speed up the de novo molecular design process by reducing labour-cost consumption. While machine learning generative models may successfully design molecules based on SMILES strings [231], they cannot effectively represent structural information about molecular structures. So, one can use GNN directly for determining chemical characteristics by analysing the topology of graphs.[232] proposed the VAE continuous embedded method for generating small molecule graphs. This initial generation mode avoids problems that may occur while generating the graphs. The maximum range on negative log-likelihood is reduced while training the model, so it is utilised only to design small compounds.[179] designed a probability specification-based method of the GCN to construct the graph in phases instead of generating instantly the whole graph gives better results than the typical way. However, there are still obstacles in generating polymer graphs.[179] suggested a conditional-graph generation model based on **graph generation** as a **Markov process** & **recursive units** there at molecular scale to optimise scalability. [122] combine the prior knowledge of the sample molecular data set to generate desired molecules. This method integrates & extends three notions of graph representation & reinforcement learning to achieve the intended aim by directing the procedure & restricting the output drug space as per basic chemical principles. The results reveal that this strategy outperforms all others in terms of maximising chemical features & restricted features under scenarios analogous to those encountered in existing chemical substances.

In contrast to the node-by-node approach of graph generation,[233]suggested linking tree self-encoding technique that generated molecular by forming connection tree structure as a practical subgraph component & joined all the sub-graphs into the full molecular graph. This model reduced the number of invalid molecular intermediate states & increased labour efficiency [234] utilised GCN to design a reactive binding model from combined data & suggested a feature prediction module that used a scoring system to find the much more favourable molecules with a particular attribute throughout the generation phase. The molecular generation approach based on graphs currently surpasses the rule-based generating mechanism. Even though the novel molecule created using the molecular graph approach scored higher on many assessment criteria, it has been challenged like the approach for building molecular graphs is now confined to 2D as 3D information of compounds needs to be addressed, which might become a priority in future.

5) **Prediction of Drug Response**
The simulation for the development of personalised medicine requires combining genomic data & drug information to predict drug response. [138] used the GCN with auto-encoder to identify the relation involving miRNA & antibiotic resistance as a semi-supervised learning problem & a mix of known miRNA expression patterns & drug structural fingerprint information was utilised to develop the **model.** Several other researchers are concentrating on the **impact** of medication treatment on the increase in the number of cells as a consequence of cell growth & division.[235] constructed the cancer cell information subnet & drug structure subnet for predicting the drug's therapeutic effect on cancer cells. Regarding the complexities of cancer causes, [236] combines genomics, biological networks, disease-gene-related data & inhibitor analysis into a **huge heterogeneous graph**. Using many graph attention propagation & convolution blocks, aggregated the network topology information & built a graphical screening framework to predict the result. [237] predict drug-induced liver damage by the structure similar to [138] but have to provide various information on dosing & drug duration as input.

6) **Drug-Drug Interaction Prediction**
A single medicine cannot address a wide range of complicated diseases. Polypharmacy is a viable strategy for combating these diseases. It entails treating patients with a cocktail of medications. However, a key disadvantage of polypharmacy is the high potential of introducing side effects resulting from drug-drug interactions. As a result, it is critical to predicting polypharmacy's adverse effects while using innovative medication when treating disorders with novel medication combinations. The polypharmacy side effect prediction task aims to assess whether the side effect occurs between two medications and their corresponding type. Computer-assisted DDI detection is the way to investigate drug combinations' medicinal characteristics.[5] estimated adverse effects between medications using a multimodal graph network. Various edges in this multimodal network depict PPI, drug-protein target, and drug interaction. Focusing on the attention mechanism used to evaluate drug similarity,[238] proposed the **multiview Drug Graph Encoder framework.** In order to effectively use the non-uniform association between diverse views,each kind of medicinal property is considered a perspective associated with acquiring attention- weights.

This multiview scheme allows getting extra-related information than the single prior view.

## C. Medical Imaging

Medical visuals are significant in therapeutic disease detection, categorisation & medication. Deep learning technologies are integrated via Computer-Aided Diagnosis for early identification & assessment of diseases. In medical imaging, like all other image-related operations, identification, classification & segmentation are primary tasks of interest. So, the graph structure is appropriate for representing image data using GNNs.

### 1) Image Segmentation

A gated graph neural network is employed for segmenting visuals & utilising directed graph learning to estimate the motion of points through coarse segmentation [[238]. In contrast, Subsequent segmentation is performed to get a smooth visual. For assessing surface data like MRI, [94] employed spectral convolution, resulting in generation-producing surface parcellation of the brain cortex. Traditional Spectral embedding can only be performed on orthogonal grid spaces. However, this approach can learn the surface features of the cortex directly. GCN block-based auto-responsive prostate contour detection approach is proposed [239] that can explain multi-scale properties & employed a training method based on the contour fit loss to retain the features for a prostate boundary. Segmentation Models in Deep learning adopt a pixel-by-pixel approach, which is computationally complex. In comparison, the **GNN approach** utilises the object's outlines for **segmentation**, which **reduces** the number of computations.

### 2) Multimodal Fusion

Together all the above findings are based on single-image analysis. However, the sole use of data from **mono-modal imaging** for disease prediction can result in a lack of accuracy in the findings. The weight-based Edge Graph Attention-network model integrates several medical imaging modalities to diagnose bipolar illnesses (FMRI or structural brain magnetic resonance imaging) [240]. Furthermore, more studies integrate medical imaging data with non-imaging variables irrelevant to disease prediction. The absence of critical information from mono-modal data may be improved & supplemented to some level via multimodal fusion. For the first time, [241] employed GCN in group-level medical applications. They introduced a population graph in which persons were considered nodes of the network & feature map of a brain image depicts the features of nodes. Phenotypic data like age and gender were integrated for illness prediction through a semi-supervised approach.[242],[243] extended the work with in-depth analytical methodologies & modelling options. After that, many researchers have opted to apply population graph approaches for disease prediction [244]. They employed a multi-level parallel GCN model, which comprised an independent learning layer enabling the distribution of weights & an attention model for harnessing the features of each multimodal data set [244] to maximise the retrieval of correlation knowledge among nodes. Inception GCN was suggested by [246] to solve the problem of inadequate extracting features caused by fixed neighbourhoods in the GCN model. **Inception GCN** considered receptive field kernels for convolution of varied dimensions & utilised two aggregating approaches to handle all the attributes acquired by the convolution kernel. In subsequent studies, the LSTM-attention method was extended to incorporate better **multimodal data** [245]. Reference [246] provided a more robust technique for **classifying autism spectrum disorders**, relying on a group of inadequately trained G-CNN used for minimising model responsiveness towards the graph design option. [243] tried to forecast brain age using a population graph, but their results need to be improved. GNNs are ideal for spotting linked patterns of brain sickness & help find the disease's process rapidly because of the morphological variations between the different parts of the brain.

## X. IMPLEMENTATION

For building deep learning models on graphs, several free Platforms or repositories have recently become accessible. The list of these libraries is in Table IIA. Several tasks involving graphs are made available to evaluate the effectiveness evaluation of various graph neural networks. Various basic datasets are listed in Table IIIA. These frequently used datasets serve as the foundation /Benchmark for such tasks. Additionally, as mentioned in Table IVA, more graph datasets are available in a broad spectrum of open-source dataset repositories. The reference to the most fully open implementations of several well-known GNN models studied in this paper covering the whole Classification of GNNs primarily from their original authors is mentioned in Table VA. It is simple to learn, evaluate, and advance various approaches to GNNs by using these open-source code implementations.

## XI. DISCUSSION & FUTURE DIRECTION

Although GNNs have demonstrated their ability to learn graph data, there are still obstacles due to the complex nature of graphs. Eight future directions for GNNs are in this section.

**Model depth:** Li et al. show that when graph convolution layers increase, all node expressions will settle to a specific location when graph convolutions drive representations of neighbouring nodes closer together [247]. It leads to whether learning graph data by going deep is still a suitable method.

**Scalability trade-off:** Increasing GNN scalability results in reducing graph completeness. A model will lose some graph information whether it utilises sampling or clustering. A future research subject might be balancing algorithm scalability with graph integrity.

**Robustness:** GNNs are subject to adversarial assaults because they are built on neural networks. In contrast to adversarial attacks involving text or images, which focus only on attributes, attacks on graphs include topological information. Several studies were published in order to attack well-established graph models. [110]; [168], as well as more robust models to defend them [110]; [168] [104]. For a complete assessment, refer [248].

**Interpretability:** For neural models, interpretability is also a significant study area. GNNs are black boxes with no explanations. It is crucial to employ GNN models in realistic conditions with appropriate explanations.

**Graph Pre-training:** Models of Neural networks require a lot of labelled data, yet getting much human-labelled data is expensive. Models are guided to learn using unlabeled data easily obtained from websites or knowledge bases using self-supervised approaches. Recent studies concentrated on graph pre-training [128] [159]; however, they all have various problem settings & focus on different features. Many open challenges exist, such as the design of pre-training activities and the efficiency of current GNN models in learning features.

**Heterogeneity:** Current GNNs are challenging to apply directly to heterogeneous graphs containing different kinds of edges & nodes, varied in the input form, such as text & pictures. As a result, new strategies for dealing with heterogeneous graphs need to be created.

**Dynamicity:** In nature, graphs are dynamic because nodes or edges inputs can alter over time. Unique graph convolution layers are needed to respond to the graph dynamicity. Although STGNNs can help with graphs' dynamicity, few investigate how to implement graph convolutions due to the presence of spatial dynamic interactions.

**Bioinformatics Methodology Related Challenges**

The methods for processing biological tasks in current GNNs should be improved for the following three application sectors of **Bio-informatics**:

**Disease Prediction**: First, most disease prediction research used broad similarity approaches. The heterogeneous graph represents detailed knowledge of RNA functional relatedness, disease semantic relatedness & multiple relationship information extracted using GNN models. However, adding multiple similarity networks to the GNN model would have increased its complexity, & a more efficient similarity evaluation paradigm is needed for novel diseases & RNA. Furthermore, more emphasis should be made on integrating node feature data inside the modelling process (like RNA structural properties & disease semantic features) to minimise over-reliance on linked data. Present GNNs are primarily designed to handle isomorphic graphs & are unable to adequately reflect the diversity of nodes & edges in heterogeneous networks [141]. As a result, a new design must be investigated to account for data characteristics in diverse bio-networks.

**Drug Discovery:** The design method of chemical networks & specification of the structure of molecular models must be revised for drug discovery. Additionally, present molecular modelling relies mainly on the 2-dimensional graph representation, with the 3-dimensional structure being ignored, which might alter molecule properties. GNN research on the chemical 3D structure might be a prospective avenue.

**Multimodal Fusion:** In medical imaging, graph neural networks have difficulties processing multimodal data. Although finding somewhat balanced data in multimodal fusion investigations is typical, more than unbalanced data is needed in practical tasks. As a result, more research into the mechanism of analysing imbalanced data in GNNs is necessary.

**Biological Data Challenges:** Many **bio-molecular** graph networks consist of sparseness & scale-free nature as their traits, i.e. many nodes also have a moderate count of connection links & just a few nodes have a large count. It reveals that a few bio-molecular nodes have a significant role in modifications of bio-molecular networks dynamically. Dropout & regularisation are two strategies that can be utilised to relieve the overfitting induced by the sparseness & scale-free feature of bio-molecular networks. During the training process, the dropout strategy randomly puts each weighted dimension to zero with a preset probability, causing the model only to update a subset of the parameters each time. Dropouts can lessen the instability when significantly less data is in the training set. The regularisation approach adds the regularisation factor to the error function to check the parameters' size. The acquisition of negative data samples is often neglected in ongoing research, leading to a need for more diversity in biological data. It is hard to train a model when no negative instances are available. Finally, since the bio-molecular network includes a lot of noise information, noise reduction processing is a great way to improve the model's performance.

## XII. CONCLUSION

Graph Neural Network models have emerged as sophisticated & valuable tools for applications of machine learning in the graph domain. This success stems from breakthroughs in expressive capability, model adaptability & training approaches of GNN. Considering the present study, we provide an in-depth analysis of graph neural networks. We present versions of GNN models categorised based on training strategies, computation units & types of graphs. Besides, we also review multiple broad frameworks & provide various conceptual cum mathematical assessments. We introduced a wide range of general applications of GNN, including a brief introduction to data sets & open-source code resources of GNN models. Then provide a thorough evaluation for applications within every case. Advances of GNN in bioinformatics are discussed from several aspects. Typical functionalities learnt via GNNs are recognised primarily on three aspects of structural information, i.e. connection-link determination, node classification & graph formation. Depending on the specific applications for varied data sets, we identify & study the associated research in three aspects: drug discovery, disease prediction & bio-medical imaging.

Eventually, future directions of utilising GNNs in general scenarios in bio-informatics investigations are highlighted. Despite GNN's outstanding results in numerous biological processes, it still needs to improve on low-quality methodology, interpretability & data processing. GNNs are indeed a practical approach that addresses several difficulties in bioinformatics inquiry. In addition, this research piece might be a significant resource for future researchers undertaking their study in this domain.

# APPENDIX

TABLE IA: OVERVIEW of ConvGNNS

Note: The absence of values ("-") in the pooling and readout layers indicates that the approach is solely testing node/edge-level activities

| Sub-class | Method | Convolution | Other Features | Read-out |
|---|---|---|---|---|
| Spectral | *Bruna et al.* | Interpolation kernel | No mutigraph | Hierarchical clustering + FC |
| | *Henaff et al* | Interpolation kernel | Constructing the graph, No mutigraph | Hierarchical clustering + FC |
| | *CayletNet* | Polynomial | No mutigraph | Hierarchical clustering + FC |
| | *GWNN* | Wavelet transform | No mutigraph | - |
| Spectral/Spatial | *ChebNet* | Polynomial | - | Hierarchical clustering |
| | *Kipf & Welling* | First-order | - | |
| Spatial | *Neural FPs* | First-order | - | Sum |
| | *PATCHY-SAN* | Polynomial + an order | A neighbor order | An order + pooling |
| | *LGCN* | First-order + an order | A neighbor order | |
| | *SortPooling* | First-order | A node order | An order + pooling |
| | *DCNN* | Polynomial diffusion | Edge features | Mean |
| | *DGCN* | First-order + diffusion | - | - |
| | *MPNNs* | First-order | A general framework | Set2set |
| | *GraphSAGE* | First-order + sampling | A general framework | - |
| | *MoNet* | First-order | A general framework | Hierarchical clustering |
| | *GNs* | First-order | A general framework | A graph representation |
| | *Kearnes et al.* | Weave module | Edge features | Fuzzy histogram |
| | *DiffPool* | Various | Differentiable pooling | Hierarchical clustering |
| | *GAT* | First-order | Attention | |
| | *GaAN* | First-order | Attention | - |
| | *HAN* | Meta-path neighbors | Attention | - |
| | *CLN* | First-order | | |
| | *PPNP* | First-order | Teleportation connections | - |
| | *JK-Nets* | Various | Jumping connections | |
| | *ECC* | First-order | Edge features | Hierarchical clustering |
| | *R-GCNs* | First-order | Edge features | |
| | *LGNN* | First-order + LINE graph | Edge features | - |
| | *PinSage* | Random walk | Neighborhood sampling | |
| | *StochasticGCN* | First-order + sampling | Neighborhood sampling | |
| | *FastGCN* | First-order + sampling | Layer-wise sampling | - |
| | *Adapt* | First-order + sampling | Layer-wise sampling | |
| | *Li et al.* | First-order | Theoretical analysis | |
| | *SGC* | Polynomial | Theoretical analysis | - |
| | *GFNN* | Polynomial | Theoretical analysis | |
| | *GIN* | First-order | Theoretical analysis | Sum + MLP |
| | *DGI* | First-order | Unsupervised training | |

TABLE IIA: POPULAR PLATFORMS/LIBRARIES FOR GRAPH COMPUTING.

| Platform | Reference |
|---|---|
| Deep Graph Library | "https://github.com/dmlc/dgl" |
| AliGraph | "https://github.com/alibaba/aligraph" |
| PyTorch Geometric | "https://github.com/rusty1s/pytorch_geometric" |
| Paddle Graph Learning Euler | "https://github.com/PaddlePaddle/PGL" |
| OpenNE | "https://github.com/thunlp/OpenNE/tree/pytorch" |
| CogDL | "https://github.com/THUDM/cogdl" |
| Plato | "https://github.com/tencent/plato" |
| GraphVite | "https://github.com/DeepGraphLearning/graphvite" |

TABLE IIIA: OTHER OPEN SOURCE DATASETS REPOSITORIES

| Repository | Introduction | Web-Links |
|---|---|---|
| Relational-Dataset Repository | To promote the development of relational-machine learning | "https://relational.fit.cvut.cz/" |
| Network Repository | A repository of scientific network-data with interactive mining and visualisation tools. | "http://networkrepository.com" |
| Open Graph Benchmark (OGB) | It specifies a set of graph machine learning data-loaders, benchmark-datasets and evaluators written in PyTorch. | "https://ogb.stanford.edu" |
| Stanford Large Network (SNAP) | The library of SNAP is designed for the purpose of investigating extensive social & information networks. | "https://snap.stanford.edu/data/" |
| Kernel Datasets | Graph-kernels benchmark-datasets | "https://ls11-www.cs.tu-dortmund.de/staff/morris/graphkerneldatasets" |

TABLE IVA: DATASETS COMMONLY USED IN TASKS RELATED TO GRAPH.

| Application Domain | Data-sets |
|---|---|
| Citation Networks | Pubmed [249] Cora [249] Citeseer [249] DBLP [187] |
| Bio-chemical Graphs | MUTAG [153] NCI-1 [222] PPI [5] D&D [250] PROTEIN [251] PTC [252] |
| Social Networks | Reddit [27] BlogCatalog [105] |
| Knowledge Graphs | FB13 [69] [69] FB15K [253] FB15K237 [252] [69] FB15K [253] WN18RR [254] |

TABLE VA: SUMMARY OF OPEN-SOURCE IMPLEMENTATIONS OF VARIOUS CLASSES OF GNN MODEL

| Category | Method | URL | Framework/Language |
|---|---|---|---|
| Graph RNNs | GGS-NNs | ``https://github.com/yujiali/ggnn'' | Torch/Lua |
| | SSE | ``https://github.com/Hanjun-Dai/steady_state_embedding'' | C |
| | You et al. | ``https://github.com/JiaxuanYou/graph-generation'' | PyTorch/Python |
| GCNs | RMGCNN | ``https://github.com/fmonti/mgcnn'' | TensorFlow/Python |
| | ChebNet | ``https://github.com/mdeff/cnn_graph'' | TensorFlow/Python |
| | Kipf&Welling | ``https://github.com/tkipf/gcn'' | TensorFlow/Python |
| | CayletNet | ``https://github.com/amoliu/CayleyNet'' | TensorFlow/Python |
| | GWNN | ``https://github.com/Eilene/GWNN'' | TensorFlow/Python |
| | Neural FPs | ``https://github.com/HIPS/neural-fingerprint'' | Python |
| | PATCHY-SAN | ``https://github.com/seiya-kumada/patchy-san'' | Python |
| | LGCN | ``https://github.com/divelab/lgcn/'' | TensorFlow/Python |
| | SortPooling | ``https://github.com/muhanzhang/DGCNN'' | Torch/Lua |
| | DCNN | ``https://github.com/jcatw/dcnn'' | Theano/Python |
| | DGCN | ``https://github.com/ZhuangCY/Coding-NN'' | Theano/Python |
| | MPNNs | ``https://github.com/brain-research/mpnn'' | TensorFlow/Python |
| | GraphSAGE | ``https://github.com/williamleif/GraphSAGE'' | TensorFlow/Python |
| | GNs | ``https://github.com/deepmind/graph_nets'' | TensorFlow/Python |
| | DiffPool | ``https://github.com/RexYing/graph-pooling'' | PyTorch/Python |
| | GAT | ``https://github.com/PetarV-/GAT'' | TensorFlow/Python |
| | GaAN | ``https://github.com/jennyzhang0215/GaAN'' | MXNet /Python |
| | HAN | ``https://github.com/Jhy1993/HAN'' | TensorFlow/Python |
| | CLN | ``https://github.com/trangptm/Column_networks'' | Keras/Python |
| | PPNP | ``https://github.com/klicperajo/ppnp'' | TensorFlow/Python |
| | JK-Nets | ``https://github.com/mori97/JKNet-dgl'' | DGL/Python |
| | ECC] | ``https://github.com/mys007/ecc'' | PyTorch/Python |
| | R-GCNs | ``https://github.com/tkipf/relational-gcn'' | Keras/Python |
| | LGNN | ``https://github.com/joanbruna/GNN_community'' | Torch/Lua |
| | StochasticGCN | ``https://github.com/thu-ml/stochastic_gcn'' | TensorFlow/Python |
| | FastGCN | ``https://github.com/matenure/FastGCN'' | TensorFlow/Python |
| | Adapt | ``https://github.com/huangwb/AS-GCN'' | TensorFlow/Python |
| | Li et al. | ``https://github.com/liqimai/gcn'' | TensorFlow/Python |
| | SGC | ``https://github.com/Tiiiger/SGC'' | PyTorch/Python |
| | GFNN | ``https://github.com/gear/gfnn'' | PyTorch/Python |
| | GIN | ``https://github.com/weihua916/powerful-gnns'' | PyTorch/Python |
| | DGI | ``https://github.com/PetarV-/DGI'' | PyTorch/Python |
| GAEs | SAE | https://github.com/quinngroup/deep-representations clustering | Keras/Python |
| | SDNE | ``https://github.com/suanrong/SDNE'' | TensorFlow/Python |
| | DNGR | ``https://github.com/ShelsonCao/DNGR'' | Matlab |
| | GC-MC | ``https://github.com/riannevdberg/gc-mc'' | TensorFlow/Python |
| | DRNE | ``https://github.com/tadpole/DRNE'' | TensorFlow/Python |
| | G2G | ``https://github.com/abojchevski/graph2gauss'' | TensorFlow/Python |
| | VGAE | ``https://github.com/tkipf/gae'' | TensorFlow/Python |
| | DVNE | ``http://nrl.thumedialab.com'' | TensorFlow/Python |
| | ARGA/ARVGA | ``https://github.com/Ruiqi-Hu/ARGA'' | TensorFlow/Python |
| | NetRA | ``https://github.com/chengw07/NetRA'' | PyTorch/Python |
| Graph RLs | GCPN | ``https://github.com/bowenliu16/rl_graph_generation'' | TensorFlow/Python |
| | MolGAN | ``https://github.com/nicola-decao/MolGAN'' | TensorFlow/Python |
| | GAM | ``https://github.com/benedekrozemberczki/GAM'' | /Pytorch /Python |
| | DeepPath | ``https://github.com/xwhan/DeepPath'' | TensorFlow/Python |
| | MINERVA | ``https://github.com/shehzaadzd/MINERVA'' | TensorFlow/Python |
| Graph adversarial methods | GraphGAN | ``https://github.com/hwwang55/GraphGAN'' | TensorFlow/Python |
| | GraphSGAN | ``https://github.com/dm-thu/GraphSGAN'' | PyTorch/Python |
| | NetGAN | ``https://github.com/danielzuegner/netgan'' | TensorFlow/Python |
| | Nettack | ``https://github.com/danielzuegner/nettack'' | TensorFlow/Python |
| | Dai et al. | ``https://github.com/Hanjun-Dai/graph_adversarial_attack'' | PyTorch/Python |
| | Zugner&Gunnemann | ``https://github.com/danielzuegner/gnn-meta-attack'' | TensorFlow/Python |
| Applications | DeepInf | ``https://github.com/xptree/DeepInf'' | PyTorch/Python |
| | Ma et al. | ``https://jianxinma.github.io/assets/disentangle-recsys-v1.zip'' | TensorFlow/Python |
| | CGCNN | ``https://github.com/txie-93/cgcnn'' | PyTorch/Python |
| | Ktena et al | ``https://github.com/sk1712/gcn_metric_learning'' | Python |
| | Decagon | ``https://github.com/mims-harvard/decagon'' | PyTorch/Python |
| | Parisot et al. | ``https://github.com/parisots/population-gcn'' | TensorFlow/Python |
| | Dutil et al. | ``https://github.com/mila-iqia/gene-graph-conv'' | PyTorch/Python |
| | Bastings et al. | ``https://github.com/bastings/neuralmonkey/tree/emnlp_gcn'' | TensorFlow/Python |
| | Neural-dep-srl | ``https://github.com/diegma/neural-dep-srl'' | Python/Therano |
| | Garcia&Bruna | ``https://github.com/vgsatorras/few-shot-gnn'' | PyTorch/Python |
| | S-RNN | ``https://github.com/asheshjain399/RNNexp'' | Therano/Python |
| | 3DGNN | ``https://github.com/xjqicuhk/3DGNN'' | Caffe/ Matlab |
| | GPNN] | ``https://github.com/SiyuanQi/gpnn'' | PyTorch/Python |
| | STGCN | ``https://github.com/VeritasYin/STGCN_IJCAI-18'' | TensorFlow/Python |
| | DCRNN | ``https://github.com/liyaguang/DCRNN'' | TensorFlow/Python |
| | Allamanis et al. | ``https://github.com/microsoft/tf-gnn-samples'' | TensorFlow/Python |
| | Li et al. | ``https://github.com/intel-isl/NPHard'' | TensorFlow/Python |
| | TSPGNN | ``https://github.com/machine-reasoning-ufrgs/TSP-GNN'' | TensorFlow/Python |
| | CommNet | ``https://github.com/facebookresearch/CommNet'' | Torch/Lua |
| | Interaction network | ``https://github.com/jaesik817/Interaction-networks_tensorflow'' | TensorFlow/Python |
| | Relation networks | ``https://github.com/kimhc6028/relational-networks'' | PyTorch/Python |
| Miscellaneous | SGCN | ``http://www.cse.msu.edu/~derrtyle/'' | PyTorch/Python |
| | DHNE | ``https://github.com/tadpole/DHNE'' | TensorFlow/Python |
| | AutoNE | ``https://github.com/tadpole/AutoNE'' | Python |
| | Gnn-explainer | ``https://github.com/RexYing/gnn-model-explainer'' | PyTorch/Python |
| | RGCN | ``https://github.com/thumanlab/nrlweb'' | TensorFlow/Python |
| | GNN-benchmark | ``https://github.com/shchur/gnn-benchmark'' | TensorFlow/Python |

*Corresponding Author:*
**Adil Mudasir Malla,** CSE.
Department of Computer Science & Engineering, Islamic University of Science & Technology; Jammu & Kashmir, India
adil.mudasir@iust.ac.in

**Asif Ali Banka,** Assistant Professor.
Department of Computer Science & Engineering, Islamic University of Science & Technology; Jammu & Kashmir, India.
asif.banka@islamicuniversity.edu.in